\documentclass{ametsocV6.1}
\usepackage{booktabs} 
\usepackage{subfig}

\title{A Diffusion-Based Framework for High-Resolution Precipitation Forecasting over CONUS}

\authors{Marina Vicens-Miquel,\aff{a,b,c}\correspondingauthor{Marina Vicens-Miquel, marina.vicensmiquel@ou.edu} 
Amy McGovern,\aff{a,b,c} 
Aaron J. Hill,\aff{a,c} 
Efi Foufoula-Georgiou,\aff{d} 
Clement Guilloteau,\aff{d} 
Samuel S. P. Shen\aff{e}}

\affiliation{\aff{a}{University of Oklahoma, School of Meteorology, Norman, Oklahoma}\\
\aff{b}{University of Oklahoma, School of Computer Science, Norman, Oklahoma}\\
\aff{c}{NSF AI Institute for Research on Trustworthy AI in Weather, Climate, and Coastal Oceanography, Norman, Oklahoma }\\
\aff{d}{University of California Irvine, Irvine, California}\\
\aff{e}{San Diego State University, San Diego, California}
}

\abstract{Accurate precipitation forecasting is essential for hydrometeorological risk management, especially for anticipating extreme rainfall that can lead to flash flooding and infrastructure damage. This study introduces a diffusion-based deep learning (DL) framework that systematically compares three residual prediction strategies differing only in their input sources: (1) a fully data-driven model using only past observations from the Multi-Radar Multi-Sensor (MRMS) system, (2) a corrective model using only forecasts from the High-Resolution Rapid Refresh (HRRR) numerical weather prediction system, and (3) a hybrid model integrating both MRMS and selected HRRR forecast variables. By evaluating these approaches under a unified setup, we provide a clearer understanding of how each data source contributes to predictive skill over the Continental United States (CONUS). Forecasts are produced at 1‑km spatial resolution, beginning with direct 1‑hour predictions and extending to 12 hours using autoregressive rollouts. Performance is evaluated using both CONUS‑wide and region-specific metrics that assess overall performance and skill at extreme rainfall thresholds. Across all lead times, our DL framework consistently outperforms the HRRR baseline in pixel-wise and spatiostatistical metrics. The hybrid model performs best at the shortest lead time, while the HRRR‑corrective model outperforms others at longer lead times, maintaining high skill through 12 hours. To assess reliability, we incorporate calibrated uncertainty quantification tailored to the residual learning setup. These gains, particularly at longer lead times, are critical for emergency preparedness, where modest increases in forecast horizon can improve decision-making. This work advances DL-based precipitation forecasting by enhancing predictive skill, reliability, and applicability across regions.}

\begin{document}

\maketitle

%
%
%
\statement
Forecasting high-resolution precipitation over large areas like the Continental United States (CONUS) is essential yet challenging due to complex spatial and temporal variability. We introduce a deep learning-based framework that produces 1 km hourly forecasts up to 12 hours in advance, improving upon the High-Resolution Rapid Refresh (HRRR) numerical weather prediction model. Our work evaluates three strategies within a unified and operationally viable setup. The proposed framework outperforms the HRRR in predictive skill at both low and high rainfall intensities, sustaining improved performance through 12-hour lead times. By incorporating interpretable uncertainty quantification, we enhance forecast transparency and trust. This approach enables more reliable, actionable predictions for emergency response, agriculture, and infrastructure planning across diverse U.S. regions.

%
%
%

\section{Introduction}
Accurately forecasting precipitation remains one of the grand challenges in atmospheric science and meteorology \citep{Beniston2013,Alexander2016,NOAA2020}. Rainfall is a complex process driven by multi-scale nonlinear interactions across space and time, influenced by topography, atmospheric dynamics, and moisture availability \citep{Smith1979,Veneziano2006,Tabari2020}. These complexities make rainfall prediction difficult, especially at fine spatial resolutions and extended lead times. Yet, timely forecasts are critical for flood risk management, water resource planning, and public safety, particularly as climate change intensifies both average and extreme rainfall patterns \citep{Dougherty2020}. Recent studies have highlighted that extreme rainfall can also lead to infrastructure failures, including dam breaches and urban flooding, resulting in significant socioeconomic and environmental impacts \citep{Hwang2024,Dharmarathne2024}.

Forecasting across large domains like the Continental United States (CONUS) introduces additional challenges, requiring models to generalize across diverse climatological regimes and complex topography \citep{Rasp2018}. Operational models such as the High-Resolution Rapid Refresh (HRRR) \citep{Dowell2022}—an implementation of the Weather Research and Forecasting model \citep{skamarock2008description}— provide hourly forecasts at 3~km resolution, which some studies argue is sufficient to capture many mesoscale processes \citep{kain2008some}. However, 3~km grids can still miss smaller-scale convective structures and localized storm dynamics that become better resolved in 1~km models \citep{weisman2023simulations}. Generating CONUS-wide forecasts at such high resolutions remains computationally expensive, but holds promise for improving representation of fine-scale extremes and storm evolution \citep{weisman2023simulations}. Moreover, skillfully forecasting both frequent, low-intensity rainfall and rare, high-impact events increases the complexity of the task, as it requires capturing a wide range of physical processes \citep{Li2012,Zhang2014}. Nevertheless, such forecasts are essential for supporting emergency management, agriculture, and infrastructure planning \citep{turner2022evaluating,merz2020impact,NRC2006}.

While machine learning (ML) methods show growing promise in precipitation forecasting, most are restricted to short lead times or limited spatial domains \citep{Sangiorgio2019, Gope2016, Kagabo2024, Vitanza2023}. Many emphasize rain/no‑rain classification or probabilistic outputs \citep{MetNet1, MetNet2, MetNet3}, which—though valuable—often require statistical expertise to interpret; a more detailed discussion of these methods and their limitations can be found in Section~\ref{sec_relat_work}. In contrast, we introduce a diffusion‑based deep learning framework that provides precipitation magnitude forecasts enhanced with integrated uncertainty quantification (UQ). To improve accessibility for decision-makers, we also propose a three-panel visualization that transparently conveys the minimum (lower bound), representative (middle), and maximum (upper bound) ensemble reconstructions—making forecast uncertainty easier to interpret and apply in decision contexts.

Our system generates hourly accumulated precipitation forecasts at 1~km resolution across the entire CONUS and extends predictions to 12 hours through autoregressive rollouts. The architecture builds on diffusion models—originally developed for generative tasks but increasingly used for spatiotemporal prediction due to their ability to produce sharp, spatially coherent outputs \citep{HoEA2020,KarrasEA2022,SongEA2020,Guilloteau2025,Gao2023,Chase2025}. Leveraging a residual-learning strategy, the model predicts the difference between inputs and targets—capturing either temporal rainfall changes or the differences in the spatial structure between HRRR and ground truth (MRMS). This design simplifies the learning task, especially at longer lead times, and improves both generalization and forecast stability.

Performance is rigorously evaluated using pixel-wise and spatiostatistical metrics over an independent test year, covering the entire CONUS and eight climatologically distinct regions. The model consistently outperforms HRRR across a range of weather regimes while maintaining robust regional skill and temporal consistency. 

Our framework also systematically compares three modeling strategies within a unified architecture: (1) a fully data-driven model using only observations, (2) an HRRR-corrective model trained only on selected HRRR forecast variables, and (3) a hybrid model integrating both HRRR and observational inputs. Previous studies often explore these strategies in isolation; we analyze them side by side using identical architectures and datasets allowing for a better understanding of how data and physical models can best be harnessed for improved forecasting.

Our approach delivers skillful, high-resolution (1 hour and 1 km) 12-hour lead time rainfall forecasts across the CONUS with improved computational scalability, interpretability, and operational utility. By combining diffusion modeling, residual learning, and UQ in a single framework, we set the stage for the next generation of deep learning-based forecasting systems.

\section{Related Work} \label{sec_relat_work}
Short-term precipitation forecasting has traditionally relied on high-resolution physical numerical weather prediction (NWP) models such as the HRRR and the Warn-on-Forecast System (WoFS). HRRR is an operational convection-allowing model with 3-km horizontal grid spacing, updated hourly with radar data assimilation for forecasts up to 18–48 hours \citep{HRRRweb,Benjamin2016}. WoFS, designed for severe weather operations, generates ensemble hourly forecasts at 1–3 km resolution with lead times of 0–6 hours and is regularly used in forecast operations at National Weather Service forecast offices around the U.S. \citep{heinselman2024warn}. While these physics-based systems effectively resolve mesoscale and convective dynamics, they are computationally intensive and require significant infrastructure for real-time deployment. Their performance may also degrade in complex terrain \citep{English2021}, limiting scalability and flexibility.

Recently, artificial intelligence (AI) and ML methods have gained traction as alternatives or complements to NWP, particularly for extreme rainfall prediction. These data-driven models offer faster inference, reduced computational costs, and improved spatiotemporal pattern recognition. Prior efforts have mostly focused on localized regions and classification tasks. For example, \citet{Sangiorgio2019} uses a deep neural network to classify extreme rainfall in Milan with a 30-minute lead time, and \citet{Gope2016} applies stacked autoencoders to forecast monsoon rainfall in India. Comparative studies have tested various ML architectures: \citet{Chkeir2023} evaluated Multilayer Perceptrons (MLPs), Long Short-Term Memory networks (LSTMs), and Encoder–Decoder LSTMs, while \citet{Kagabo2024} benchmarked LSTMs, Convolutional Neural Networks (CNNs), and Gated Recurrent Units (GRUs) in Rwanda. Other strategies such as multimodal integration and spatial clustering are applied to enhance performance in extreme rainfall detection, e.g., \citet{Vitanza2023} in Sicily. In the U.S., \citet{Hill_Schumacher2021}, \citet{Loken2019}, and \citet{JamesDissertation} use random forests for excessive rainfall prediction over CONUS subdomains. However, most studies emphasize binary outcomes (e.g., rain/no-rain or threshold exceedance) rather than generating high-resolution, precipitation magnitude forecasts.

To address these limitations, recent work has turned to more expressive models, particularly diffusion-based generative frameworks. These models produce sharper spatial outputs and better capture uncertainty than classification-based approaches. Notable studies include \citet{Guilloteau2025}, who generates probabilistic ensembles from satellite inputs; \citet{Chase2025}, who applies score-based diffusion to nowcast GOES imagery; and \citet{Gao2023} and \citet{Asperti2025}, who introduces latent diffusion approaches for precipitation nowcasting. Yet, these works are confined to very short lead times (typically $\leq$ 3 hours), highlighting the challenge of extending diffusion models for longer-range forecasting—a gap this work addresses directly. 

Google DeepMind’s MetNet series represents a prominent line of AI-driven precipitation forecasting. MetNet-1 \citep{MetNet1} introduces a convolutional neural network (CNN) with dilated convolutions and attention mechanisms to predict precipitation probabilities up to 8~hours ahead using satellite and radar data. MetNet-2 \citep{MetNet2} increases both the spatial resolution and model depth, extending forecasts to 12~hours. MetNet-3 \citep{MetNet3} further advanced the architecture by incorporating temporal fusion, sparse encoders, and deep ensembles to achieve lead times of up to 24~hours. All versions generate probabilistic forecasts for predefined precipitation thresholds.

\section{Datasets}
This study utilizes hourly data from the MRMS system \citep{Zhang2016} and the HRRR model \citep{Dowell2022}. Specifically, we employ MRMS Quantitative Precipitation Estimates (QPE) at hourly 1 km grid and accumulated HRRR hourly precipitation forecasts at 3 km, covering lead times from 1 to 12 hours. To ensure temporal consistency and avoid biases from versioning changes in NOAA’s MRMS and HRRR products, we restrict our dataset to a stable period between 01 March 2021 and 28 February 2025, during which product formats remained unchanged. Further, due to differences in spatial resolution and grid structure—projected for MRMS and lat-lon for HRRR—preprocessing is required to align both datasets for training.

\subsection{Data Preprocessing}
Our preprocessing pipeline addresses spatial alignment, missing data, and normalization to prepare the datasets for training. To ensure spatial compatibility, HRRR fields are regridded to the 1~km MRMS grid using xarray \citep{xarray_hoyer2017} and xESMF \citep{xESMF_zhuang_2018_1134366}, with bilinear interpolation applied after cropping to the target domain and generating matching latitude–longitude arrays. In terms of temporal continuity, both datasets exhibit occasional missing data, including a full month gap in MRMS during June 2021. Despite these discontinuities, we maintain temporal alignment and date consistency, skipping periods with missing data. Such gaps are common in environmental datasets due to sensor outages or quality control filtering. Finally, to improve the training stability and performance of the diffusion model, we apply z-score standardization, which aligns with the unit-variance assumptions of Gaussian noise in diffusion models \citep{HoEA2020,KarrasEA2022}.

\subsection{Inputs}
Identifying the most effective input configuration is essential for high-resolution precipitation forecasting. To this end, we conduct extensive ablation studies evaluating a broad range of HRRR variables and observational inputs (see Appendix~\ref{append_inputs_evaluated}). The results show that both observational and HRRR forecast-based predictors contribute complementary information — observational inputs improve short-term accuracy, while HRRR forecasts enhance spatial coherence and storm evolution representation.

These findings motivate the development of three distinct modeling configurations, each emphasizing a different input philosophy: purely data-driven (i.e., MRMS only), HRRR-corrective, and hybrid (see Section~\ref{sec_dataset_creation}). For all configurations, auxiliary variables such as latitude, longitude, and cyclic temporal encodings are included to provide spatial and seasonal context.

For example, the 1-hour lead time hybrid configuration combines both observational and selected HRRR forecast variables as follows:
\[
\begin{gathered}
\text{MRMS QPE}(t\!-\!2),\ \text{MRMS QPE}(t\!-\!1),\ \text{MRMS QPE}(t), \\
\text{HRRR f01 1-h Accumulated Total Precipitation}(t), \\
 \text{HRRR f02 1-h Accumulated Total Precipitation}(t), \\
 \text{Latitude},\ \text{Longitude},\ \text{Temporal Encoding}(t).
\end{gathered}
\]

This configuration leverages both recent observations and short-range forecasts to provide temporal continuity and spatial consistency. Prior MRMS fields capture recent rainfall evolution \citep{Chase2025,Shi2015ConvLSTM}, while HRRR forecasts at $t$ and $t+1$ provide information about atmospheric conditions driving storm progression.

Here, MRMS QPE$(t-i)$ denotes the $i$-hour-previous accumulated hourly precipitation from MRMS, and HRRR f$L$$(t)$ denotes the $L$-hour lead-time accumulated forecast from HRRR.

\subsection{Dataset Creation}\label{sec_dataset_creation}
To enable a systematic comparison of modeling strategies, we construct three distinct hourly input datasets, each tailored to one of the following approaches: (a) a purely observation-driven model (Data-Driven), (b) a forecast-corrective model using HRRR outputs (HRRR-Corrective), and (c) a hybrid model combining both data sources (Hybrid). These datasets are designed based on the input analysis and ablation findings from the previous section.

Table~\ref{tab_dataset_description} summarizes the inputs used for each configuration. To enhance spatial awareness and capture diurnal and seasonal variability across the CONUS, all datasets included auxiliary contextual inputs: latitude, longitude, and cyclic temporal encodings.

\begin{table}[t]
\caption{Description of the three input datasets used for model training for 1-hour lead time precipitation prediction at time step ~$t$.}
\label{tab_dataset_description}
\begin{center}
\begin{tabular}{p{2.3cm}p{3.4cm}p{6.0cm}p{2.6cm}p{0.0cm}}
\hline\hline
\centering \textit{Dataset} & \centering \textit{Observation-Based Inputs} & \centering \textit{Forecast-Based Inputs} & \centering \textit{Auxiliary Inputs} & \\
\hline
\centering \textit{Data-Driven} & \centering MRMS\((t{-}2)\), MRMS\((t{-}1)\), MRMS\((t)\) & \centering -- & \centering Latitude, Longitude, Temporal Encoding\((t)\) & \\
\hline
\centering \textit{HRRR-Corrective} & \centering -- & \centering HRRR f01(t{-}3), HRRR f01(t{-}2), HRRR f01(t{-}1), HRRR f01(t), HRRR f02(t) & \centering Latitude, Longitude, Temporal Encoding\((t)\) & \\
\hline
\centering \textit{Hybrid} & \centering MRMS\((t{-}2)\), MRMS\((t{-}1)\), MRMS\((t)\) & \centering HRRR f01(t), HRRR f02(t) & \centering Latitude, Longitude, Temporal Encoding\((t)\) & \\
\hline
\end{tabular}
\end{center}
\end{table}

\subsubsection{Training and Testing Datasets}
\begin{figure}[t]
\begin{center}
 \noindent\includegraphics[width=30pc,angle=0]{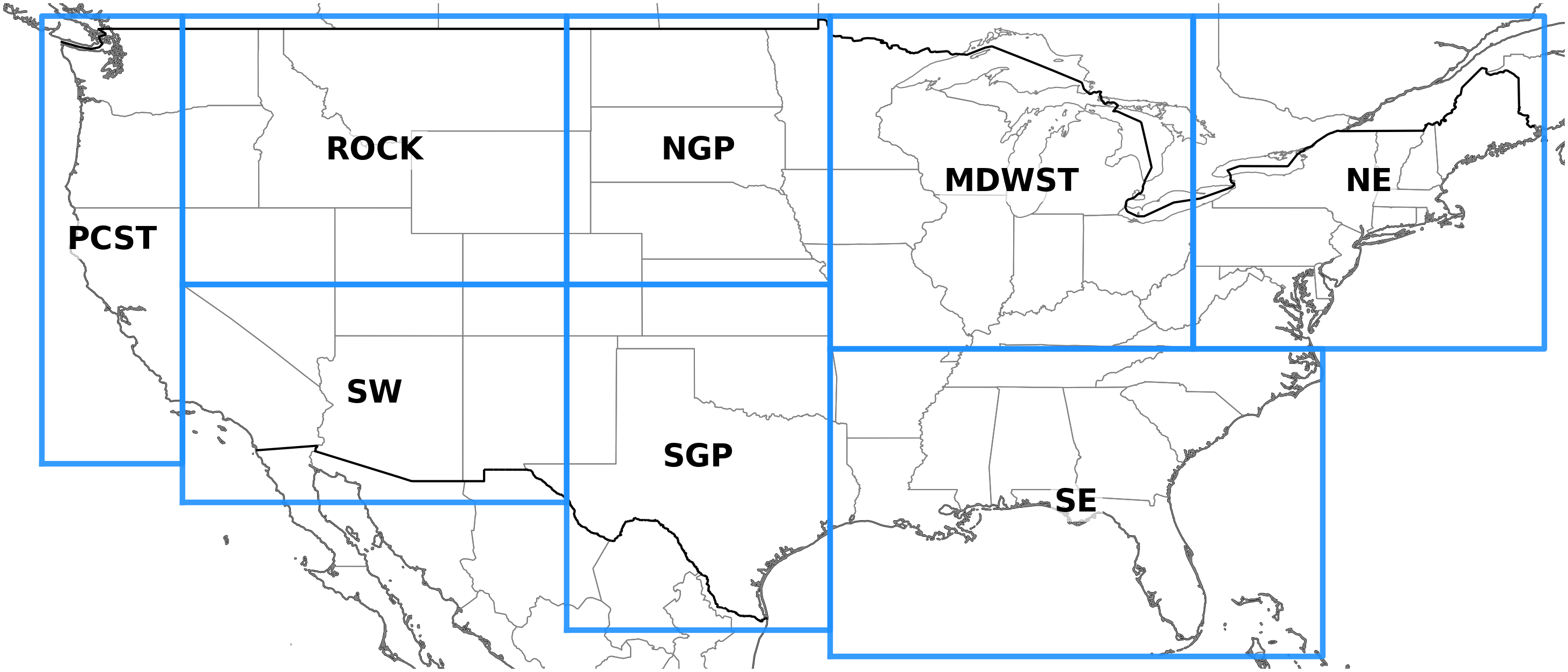}\\
 \caption{Regional divisions across the CONUS used for evaluation. These eight regions are adopted from \cite{Hill_Schumacher2021} to reflect climatological variability in convective hazards and support region-specific analysis of model performance.}
 \label{fig_regions}
\end{center}
\end{figure}

The training dataset consists of approximately 30,000 tiles of size 512×512 km, randomly sampled across the CONUS domain over a three-year period spanning March~2021 to February~2024. To ensure the inclusion of diverse and meteorologically relevant precipitation scenarios, each tile is retained only if it satisfies predefined rainfall intensity or coverage criteria. Regional balance is also enforced to achieve comparable representation across distinct climatological regimes. These regions follow the geographical delineations defined by \citet{Hill_Schumacher2021} and are strategically chosen based on rainfall and convection climatologies across the CONUS (Fig.~\ref{fig_regions}). Further details on the rainfall thresholds, regional balance, and sampling methodology are provided in Appendix~\ref{appendix_tile_sampling}.

For evaluation, we generate a CONUS-wide testing dataset composed of 512×512~km tiles with 50~km overlap, covering the period March~2024 to February~2025. To enable both monthly and regional analyses, performance is assessed independently for each month. Due to computational constraints, only two initialization times are selected per day; however, for each initialization, forecasts are produced for all 12 lead times (1–12~h). This approach effectively provides coverage across all hours of the day while maintaining a feasible computational load. Appendix~\ref{appendix_testing_strategy} details the temporal sampling strategy used to select the testing hours.

\section{Methodology} \label{sec_methodology}
This section presents our end-to-end framework for high-resolution precipitation forecasting across the CONUS. It begins by describing the residual learning strategy (Section~\ref{sec_error_corr_and_res_learn}), followed by the model architecture (Section~\ref{sec_architecture}), the uncertainty-aware loss function (Section~\ref{sec_loss}), the autoregressive rollout procedure (Section~\ref{sec_rollouts}), and our uncertainty quantification approach (Section~\ref{sec_uq}). Additional implementation details, including hyperparameter configurations and computational resources, are provided in Appendix~\ref{append_hyperparams}.

\subsection{Residual Learning and HRRR Error Correction Strategy} 
\label{sec_error_corr_and_res_learn}
Accurately forecasting high‑resolution precipitation is difficult due to its localized, nonlinear nature and the large dynamic range of rainfall values. To improve numerical stability, optimization, and multi‑step rollout behavior, we formulate the prediction task in terms of ``residuals" rather than direct rainfall amounts. Depending on the model configuration, the network learns either short‑term observational changes or corrections to HRRR forecasts.

For the observation‑driven (Data-Driven) configuration, the model predicts the residual between consecutive MRMS observations:
\[
\mathrm{Target}(t{+}1) = \mathrm{MRMS}(t{+}1) - \mathrm{MRMS}(t).
\]
This focuses learning on local rainfall evolution instead of full‑field precipitation magnitudes.

When HRRR forecasts are included (HRRR‑Corrective and Hybrid models), the target represents the forecast error at lead time $L$, where $\mathrm{HRRR\ fL}(t)$ denotes the HRRR forecast initialized at time $t$ and valid at $t{+}L$:
\[
\mathrm{Target}(t{+}L) = \mathrm{MRMS}(t{+}L) - \mathrm{HRRR\ fL}(t).
\]

Overall, the residual learning strategy (i) narrows the prediction range for easier optimization, (ii) reduces error accumulation during multi‑step rollouts, (iii) improves representation of persistent and extreme precipitation, and (iv) provides an operationally practical framework for real‑time HRRR error correction \citep{mardani2025residual}.

\subsection{Model Architecture} \label{sec_architecture}
\begin{figure}[t]
\begin{center}
 \noindent\includegraphics[width=\textwidth]{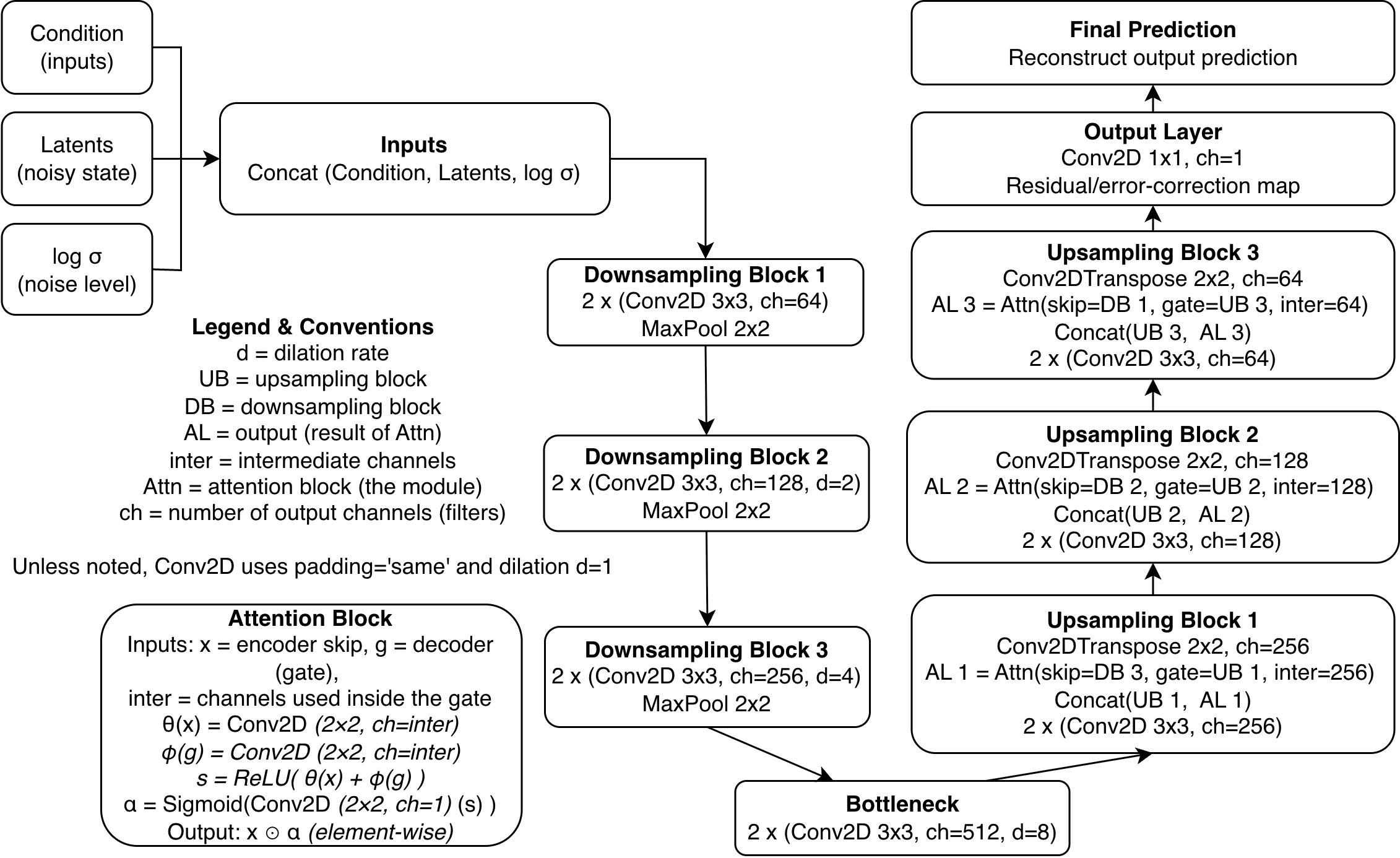}\\
 \caption{Overview of the proposed diffusion-based residual forecasting architecture. 
 A dilated attention U-Net denoiser is trained under EDM preconditioning \citep{KarrasEA2022}.
 Inputs concatenate HRRR predictors (condition), a Gaussian latent field scaled by the sampled noise level $\sigma$, and $\log \sigma$. 
 The network predicts a single-channel residual correction map (data prediction, $x_{0}$-prediction).}
 \label{fig_architecture}
\end{center}
\end{figure}

Figure~\ref{fig_architecture} illustrates the architecture of our diffusion-based residual forecasting model. It follows a dilated attention U-Net backbone wrapped within the elucidated diffusion model (EDM) preconditioning framework proposed by \citet{KarrasEA2022}. Inputs consist of HRRR meteorological predictors (the condition), Gaussian latent noise scaled by the noise level $\sigma$, and the logarithm of the noise level, $\log\sigma$. Here, $\sigma$ represents the standard deviation of the noise distribution that controls the perturbation strength applied to the input during diffusion steps. These inputs are concatenated along the channel axis and passed through an encoder–decoder network with three downsampling blocks (64, 128, 256 filters), a dilated convolutional bottleneck (512 filters, rates 1–8), and a decoder with transposed convolutions and additive attention gates. Finally, a $1 \times 1$ convolution is applied to reduce the feature channels to a single output, producing the residual (error-correction) map.

We use EDM preconditioning to stabilize training across noise levels. The wrapper introduces scale-aware coefficients ($c_{\text{skip}}, c_{\text{in}}, c_{\text{out}}$) to balance noisy and clean components, enabling robust learning across the full diffusion range. Rather than predicting the added noise ($\varepsilon$), as in conventional diffusion setups, we train the model to directly predict the clean residual field—commonly referred to as the $x_0$ (data prediction) formulation—allowing recovery of the original signal at each denoising step. At each step, we sample $\sigma$, create a noisy input, and supervise the model’s prediction of the residual field using a hybrid, $\sigma$-aware loss (Section~\ref{sec_loss}).

Data prediction aligns the model output with the true variable of interest—residual rainfall—ensuring consistency during rollout and direct supervision of both typical and extreme events. Unlike $\varepsilon$-prediction, which denoises in latent space and requires a transformation to the physical domain \citep{HoEA2020,KarrasEA2022,yu2024unmasking}, $x_0$-prediction keeps all intermediate states interpretable and consistent with the residual dynamics \citep{KarrasEA2022,meijer2024rise}. This is particularly valuable in operational settings, where accurate representation of extremes is critical.

\subsection{Loss Function} \label{sec_loss}
To achieve accurate and uncertainty-aware residual predictions across precipitation regimes, we introduce a custom loss function, HybridSigmaLoss, tailored for our residual learning task. Since the model predicts residuals—either MRMS change or HRRR forecast error—the target values are centered near zero and are often small in magnitude. Moreover, approximately 75\% of the pixels in our dataset correspond to no-rain conditions, requiring a loss function that balances sensitivity to subtle deviations with responsiveness to impactful events.

HybridSigmaLoss combines two components:
\begin{equation}
\mathrm{HybridSigmaLoss} = \alpha \cdot \mathrm{Scaled\ MAE} + (1 - \alpha) \cdot \mathrm{Weighted\ MAE}
\label{eq_loss}
\end{equation}
with $\alpha = 0.8$ controlling the balance between calibrated uncertainty and rainfall-intensity sensitivity. This value is selected through hyperparameter tuning, as values closer to 1 place greater emphasis on the Scaled MAE term, prioritizing the diffusion model’s uncertainty calibration over magnitude error.

Scaled MAE incorporates predictive uncertainty via:
\begin{equation}
\textit{Scaled MAE} = \frac{\left| y_{\text{pred}} - y_{\text{true}} \right|}{\sigma + \varepsilon}
\label{eq_scaled_mae}
\end{equation}
where $\varepsilon = 10^{-6}$ prevents division by zero. This term penalizes overconfident errors and encourages the model to be cautious where uncertainty is high, promoting well-calibrated forecasts.

Weighted MAE applies higher penalties to errors in rainfall ranges. We use a smooth sigmoid-based weighting curve:
\begin{equation}
\begin{aligned}
w =\ & \text{sigmoid}((y - 0.015) \cdot 150) \cdot 3.5 \\
&+ \text{sigmoid}((y - 0.08) \cdot 50) \cdot 5.0 \\
&+ \text{sigmoid}((y - 0.25) \cdot 20) \cdot 6.0 \\
&+ \text{sigmoid}((y - 0.5) \cdot 10) \cdot 7.0 \\
&+ \left(1 - \text{sigmoid}((y - 0.015) \cdot 150)\right) \cdot 0.6
\end{aligned}
\label{eq_weighted_mae_weights}
\end{equation}
yielding:
\begin{equation}
\textit{Weighted MAE} = w \cdot \left| y_{\text{pred}} - y_{\text{true}} \right|
\label{eq_weighted_mae}
\end{equation}

The thresholds and scaling factors in Equation \ref{eq_weighted_mae_weights} are determined through hyperparameter tuning. This process optimizes the weighting curve to balance sensitivity to light rain and high-intensity errors while maintaining numerical stability and differentiability. Additional implementation details and the corresponding weighting curve visualization are provided in Appendix~\ref{appendix_weighted_mae_curve}.

Together, these choices guide the model to produce well-calibrated predictions with strong performance across the full rainfall spectrum. Although the model predicts continuous rainfall magnitudes rather than categorical outputs, we also evaluate its skill in distinguishing rain versus no-rain conditions, as well as in capturing high-intensity and extreme events across multiple percentiles and rainfall ranges. By integrating uncertainty-awareness and dynamic weighting, HybridSigmaLoss mitigates the averaging effects that often overlook extremes and helps the model learn meaningful residual corrections across diverse rainfall regimes and lead times.

\subsection{Autoregressive Rollouts} \label{sec_rollouts}
We adopt an autoregressive rollout strategy, where the model is trained to predict residuals corresponding to 1-hour forecasts, and forecasts are extended to 12 hours by iteratively feeding previous predictions back into the input stack.

At each inference step, we:
\begin{itemize}
    \item Update the observations by replacing $\mathrm{MRMS}(t)$ with the model’s prediction $y_{\text{pred}}(t{+}1)$, and shift all past observations forward by one hour.
    \item For the Hybrid model, in addition to the previous step, we also update the HRRR input by replacing it with the forecast corresponding to the next HRRR lead time.
    \item For the HRRR-Corrective model, we keep the prior HRRR forecasts associated with past time steps unchanged, while updating only the HRRR forecast to the next lead time.
\end{itemize}

The HRRR-Corrective model exhibits sensitivity when trained on one set of HRRR forecasts but evaluated using another, likely because different lead times are generated by distinct model configurations — each introducing its own systematic and cross-lead-time biases. To mitigate this, we preserved prior HRRR forecasts during inference to maintain temporal consistency, which helped stabilize the model’s predictive behavior. We also enhance stability during training by including both HRRR f01 and f02 lead times, with f02 left unnormalized to help the model capture magnitude relationships across forecast ranges. During inference, rather than using a mismatched pair (f06 and f07) for the 6-hour forecast—both unseen during training—the model instead received the normalized and unnormalized versions of f07. This adjustment preserves amplitude consistency and prevents the compounding of unfamiliar forecast biases, since the model had not learned either the individual or joint bias characteristics of f06 and f07. Using a forecast source (f06) in dual normalized forms improved stability and yielded more reliable predictions, particularly at longer lead times. Empirically, this strategy reduces the model’s sensitivity to HRRR lead-time variability and results in higher predictive skill and more spatially coherent rainfall structures.

\subsection{Uncertainty Quantification} \label{sec_uq}
In environmental sciences and many other applied domains, UQ plays a critical role—especially in the context of operational forecasting. While deterministic models offer a single best-guess prediction, UQ provides additional insight into the confidence and reliability of those forecasts. This is essential for risk-aware decision-making, where knowing the likelihood and spread of possible outcomes can be just as important as the prediction itself.

A common approach to UQ in diffusion-based models involves sampling the latent noise multiple times to generate an ensemble of predictions, from which spread metrics (e.g., standard deviation or percentiles) can be derived. However, this strategy proves ineffective in our HRRR-Corrective setting due to two key limitations: (1) limited ensemble spread caused by the residual-based prediction structure, and (2) the high computational cost of repeated diffusion sampling.

The first limitation stems from the structure of our HRRR-Corrective model, which predicts residuals that are added to the HRRR forecast. Because these residuals represent relatively small corrections to a dominant input field, the ensemble predictions tend to be very similar, regardless of the latent noise sampled. This results in low ensemble spread and overconfident forecasts that underestimate uncertainty and fail to reflect meaningful variability. The second limitation concerns computational efficiency. Each ensemble member requires running the full denoising process, which is computationally expensive—especially when applied across the entire CONUS domain and for multi-step rollouts. Given the low added value and high cost, this approach is not used.

Instead, we develop a physics-informed alternative that accounts for known temporal uncertainties in the HRRR forecasts. Specifically, HRRR predictions often exhibit time-lag errors, where the forecasted precipitation field aligns best with observations shifted by one forecast hour earlier or later. To incorporate this into our uncertainty modeling, we modified the final reconstruction step of the HRRR-Corrective model, where the residual is added to the HRRR field, by generating a three-member ensemble based on lead-time offsets.

Recall that the final output at time \( t+L \) is computed as:
\[
\text{Prediction}(t+L) = \text{HRRR fL}(t) + \text{Residual}(t+L)
\]
where $L$ denotes the forecast lead time (in hours) relative to the initialization time $t$.

To capture temporal uncertainty—defined as variability in the timing of predicted rainfall relative to observations—we generate three forecasts per lead time:

\begin{enumerate}
    \item Early scenario: \( \text{HRRR fL-1}(t) + \text{Residual}(t+L) \)
    \item On-time scenario: \( \text{HRRR fL}(t) + \text{Residual}(t+L) \)
    \item Delayed scenario: \( \text{HRRR fL+1}(t) + \text{Residual}(t+L) \)
\end{enumerate}

Each version of the selected HRRR forecast variable represents a plausible alternative where the timing of the forecast is slightly offset. From these three variants, we construct pixel-wise uncertainty bounds by taking the minimum and maximum predicted rainfall values across the three scenarios. This yields a spatially resolved lower and upper bound, with the on-time scenario serving as the middle prediction. Specifically:

\begin{enumerate}
    \item Lower bound: pixel-wise min(early, on-time, delayed scenarios)
    \item Middle: on-time scenario
    \item Upper bound: pixel-wise max(early, on-time, delayed scenarios)
\end{enumerate}

This approach leverages known temporal biases in HRRR and provides a physically interpretable estimate of temporal uncertainty without requiring repeated diffusion sampling. Although it does not explicitly account for spatial uncertainty, the resulting bounds correspond to plausible early, on-time, or delayed forecast scenarios, aligning well with operational needs. Future work could extend this framework to jointly model temporal and spatial displacement errors in convective systems.

\section{Results and Discussion}
This section presents a comprehensive evaluation of model performance across the three configurations—Data-Driven, HRRR-Corrective, and Hybrid—benchmarked against the HRRR forecast. All predictions are compared against MRMS QPE, which serves as the ground truth. Evaluation is conducted using pixel-wise and metrics of spatio-temporal structure across independent months, lead times, and both CONUS-wide and regional scales to assess predictive skill and added value.

To ensure robust and multi-perspective evaluation, we employ a diverse set of performance metrics that capture intensity error, temporal dynamics, and spatial agreement, as well as diagnostic measures that characterize multi-scale structural behavior. Performance metrics include mean absolute error (MAE), probability of detection (POD), critical success index (CSI), and fraction skill score (FSS) for pixel-wise and binary threshold-based assessments. For brevity, we show FSS results computed with a 27 × 27 grid-cell neighbourhood. In addition, structural diagnostics such as the two-dimensional Fourier power spectrum, spectral coherence with the MRMS ground-truth field, and rainfall distribution are used to qualitatively assess scale-dependent spatial organization. Metric definitions and formulas are provided in Appendix~\ref{append_evaluation_metrics}. Binary metrics are computed using different thresholds per region (Table~\ref{tab:region_thresholds} in Appendix~\ref{append_evaluation_metrics} lists the specific values).

\subsection{CONUS Pixel-Wise Evaluation Across Lead Times 1--12 Hours}
\begin{figure}[t]
\begin{center}
 \noindent\includegraphics[width=\linewidth,angle=0]{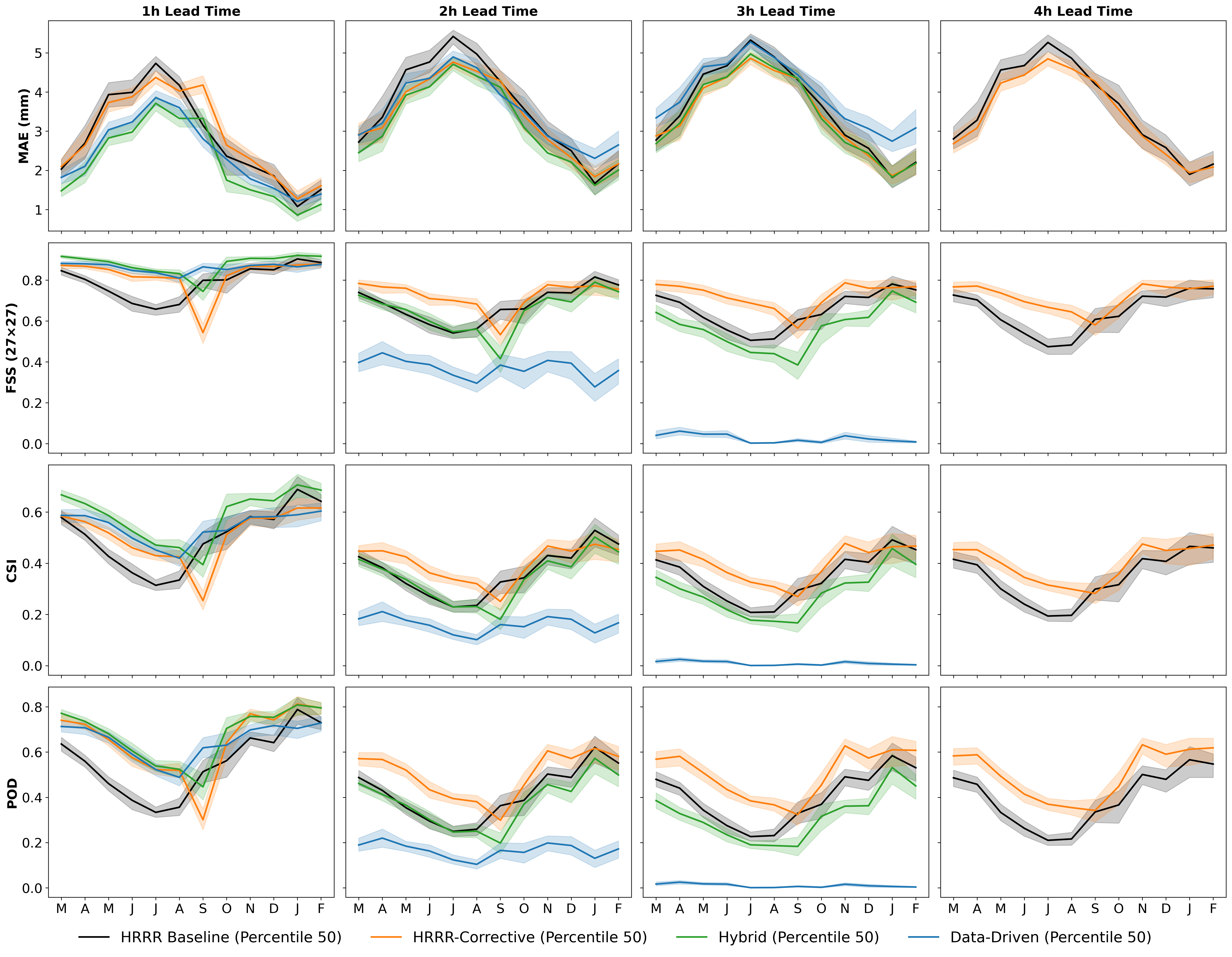}\\
 \caption{Pixel-wise evaluation metrics across independent testing months for lead times from 1h to 4h, covering the period from March 2023 to February 2024. Each letter on the x-axis corresponds to the first letter of the month (e.g., M = March, A = April, etc.). For lead times 1h to 3h, we compare the performance of the three proposed models: Data-Driven, HRRR-Corrective, and Hybrid—against the HRRR. For the 4h lead time, we focus solely on the HRRR-Corrective model, as it demonstrated the most consistent performance across earlier lead times. Shaded regions represent 95\% confidence intervals. Metrics include MAE, FSS (27×27), CSI, and POD.}
 \label{fig_pixel_eval_1to4}
\end{center}
\end{figure}

\begin{figure}[t]
\begin{center}
 \noindent\includegraphics[width=\linewidth,angle=0]{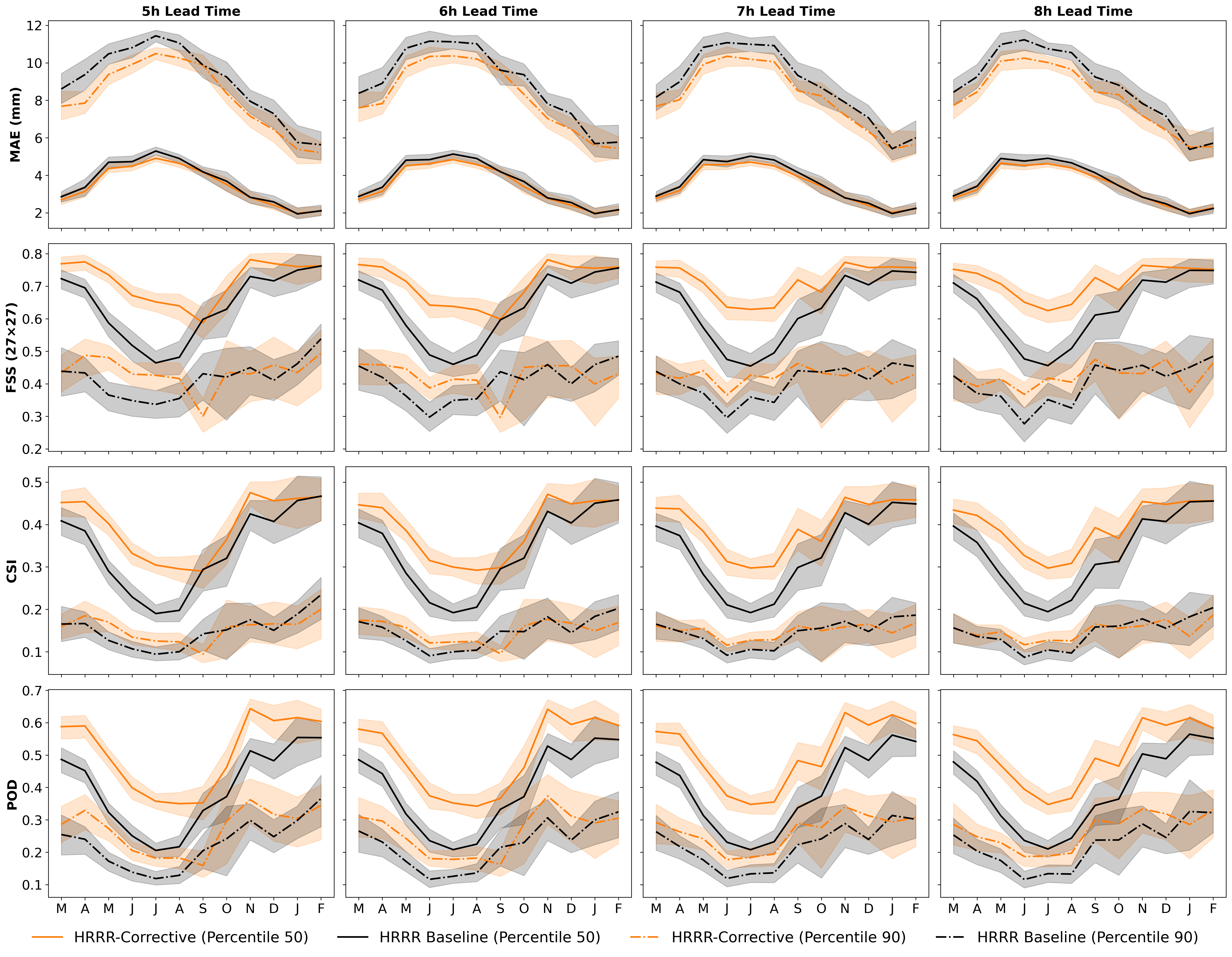}\\
 \caption{Similar to Figure \ref{fig_pixel_eval_1to4}, but for lead times from 5 h to 8 h. Results show the best-performing longer-lead-time model (HRRR-Corrective) compared against the HRRR baseline at the 50th and 90th percentiles. Shaded regions denote 95\% confidence intervals.}
 \label{fig_pixel_eval_5to8}
\end{center}
\end{figure}

\begin{figure}[t]
\begin{center}
 \noindent\includegraphics[width=\linewidth,angle=0]{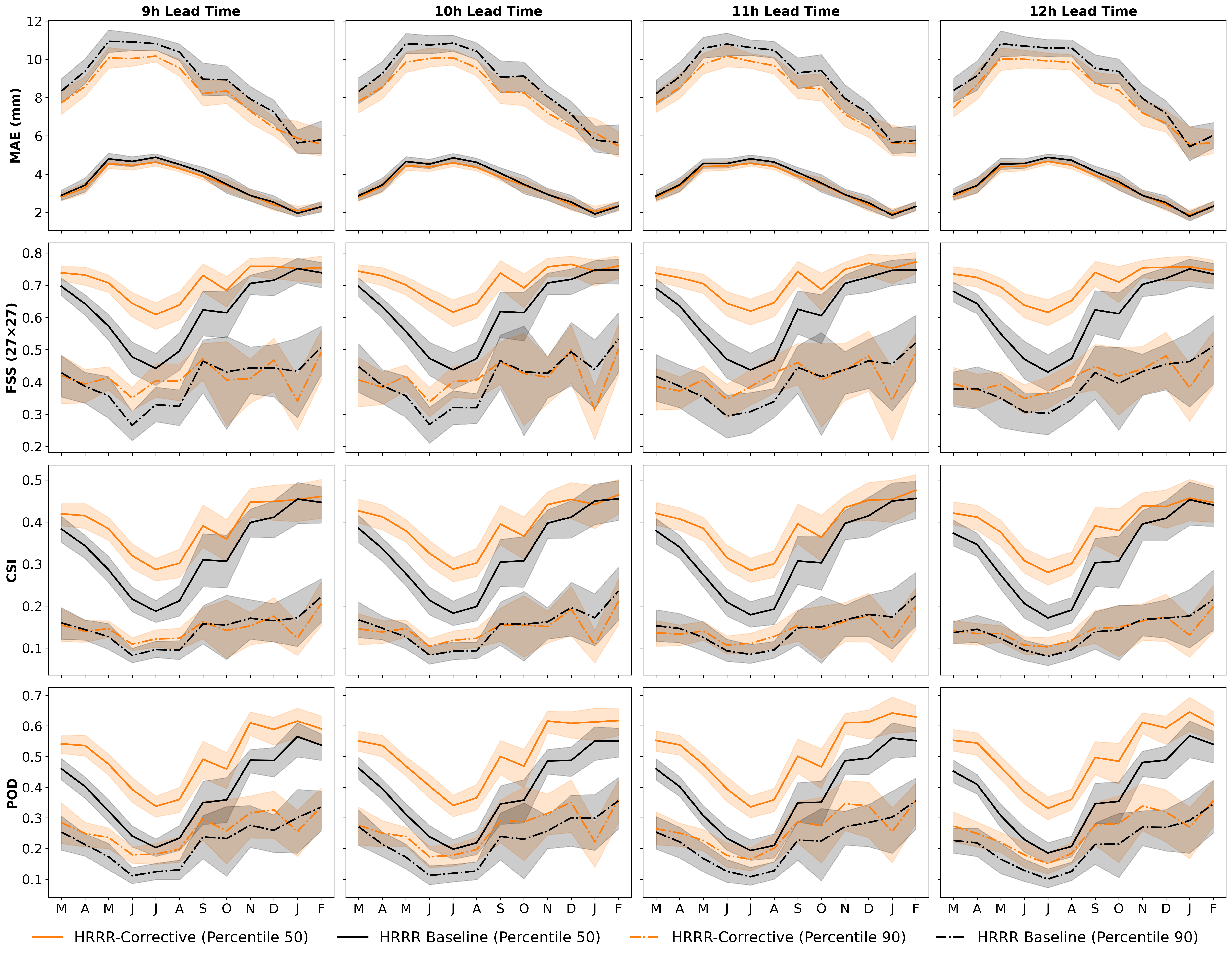}\\
 \caption{Same as Figure \ref{fig_pixel_eval_5to8}, but for lead times from 9 h to 12 h.}
 \label{pixel_eval_9to12h}
\end{center}
\end{figure}

Figures~\ref{fig_pixel_eval_1to4}, \ref{fig_pixel_eval_5to8}, and \ref{pixel_eval_9to12h} present the CONUS-wide pixel-wise evaluation of the models (Data-Driven, HRRR-Corrective, and Hybrid), benchmarked against the HRRR forecast. Results are shown for rainfall intensities corresponding to the 50th and 90th percentiles of the MRMS precipitation climatology derived from the four-year dataset used in this study. The 50th percentile represents typical rainfall conditions, while the 90th percentile captures performance under extreme precipitation events. All metrics are computed after removing zero-rainfall pixels to ensure that evaluations reflect model skill in predicting actual rainfall rather than being dominated by large dry regions. The metrics also include 95\% confidence intervals, computed using bootstrapping with n = 1,000, to quantify the uncertainty of each metric. 

At the 1-hour lead time, all AI-based models generally outperform the HRRR across most metrics and months, with the Hybrid configuration showing the strongest and most consistent improvements overall. However, the HRRR-Corrective model exhibits intermittent periods of reduced performance, particularly during late-spring and early-summer months (e.g., May–June), as shown in Fig.~\ref{fig_pixel_eval_1to4}. These fluctuations likely reflect variability in HRRR forecast skill during convective seasons, which directly affects the corrective model’s input quality.

At 2h, performance decreases for all models due to autoregressive error propagation. The Data-Driven model exhibits the largest degradation, which is consistent with expectations for architectures trained solely on observational inputs—these models depend on recent rainfall patterns and therefore cannot anticipate the initiation of new convective systems in the absence of dynamical predictors. The Hybrid model shows moderate degradation, likely due to discrepancies between the HRRR fields used during training and those provided during rollout. In contrast, the HRRR-Corrective model remains the most stable and delivers the best performance among all configurations, consistently surpassing the HRRR. Its robustness stems from the input-update strategy, which replaces HRRR fields sequentially rather than feeding back predicted precipitation, thereby preventing autoregressive error accumulation.

At 3h, the Data-Driven model continues to degrade, while the Hybrid model shows only a modest decline and remains competitive in MAE but inferior in categorical metrics (FSS, CSI, POD) compared to the HRRR model. The HRRR-Corrective model maintains high stability and outperforms HRRR across all metrics, demonstrating strong mid-range forecasting capability.

From 5 h to 12 h lead times, we focus our evaluation on the HRRR-Corrective model only, as it is the configuration that consistently outperforms HRRR at longer horizons. For 5h--8h lead times, both HRRR and the DL model show little degradation in performance, consistent with the fact that rollout does not use predicted precipitation and instead replaces HRRR fields sequentially (f01 $\rightarrow$ f02 $\rightarrow$ f03, etc.). Across these lead times, the HRRR-Corrective model achieves consistently lower MAE and higher spatial skill than HRRR for both typical (50th percentile) and extreme (90th percentile) events. The improvements are statistically significant across all metrics for most months at the 50th percentile, as indicated by non-overlapping 95\% confidence intervals. At the 90th percentile, differences are significant for MAE in most months, partially significant for POD (roughly half the months), and alternate between likely significant and comparable performance for the remaining metrics.

From 9–12 h lead times, the HRRR-Corrective model continues to outperform the HRRR. Improvements are particularly evident during the summer months, when HRRR performance typically declines—highlighting the robustness of the DL model under more challenging convective conditions. The HRRR-Corrective model remains significantly better for most months across MAE, FSS, and POD at the 50th percentile, while CSI shows likely significant improvements for some months across all lead times. At the 90th percentile, performance differences narrow, with both models showing similar skill for extreme events. Although the current analysis does not extend beyond 12h, the results suggest that the model could potentially maintain skill advantages at longer lead times as well.

A bigger picture analysis highlights pronounced seasonal variability across all models and metrics, especially during the winter and early spring months (January–April, November–December), yielding the highest performance, while warm-season months (May–September) exhibit significantly lower skill. This pattern reflects the relatively higher predictability of stratiform winter precipitation versus the localized, rapidly evolving convective storms that dominate summer months \citep{walser2004predictability,jeworrek2021wrf}. These warm-season regimes pose inherent challenges for both physical and DL models and may require higher-resolution inputs or convective-awareness strategies to improve performance \citep{li2024statistical,balmaseda2021noaa}.

\subsection{CONUS Spatiostatistical Evaluation Across Lead Times 1–12 Hours}
\begin{figure}[t]
\begin{center}
 \noindent\includegraphics[width=\linewidth,angle=0]{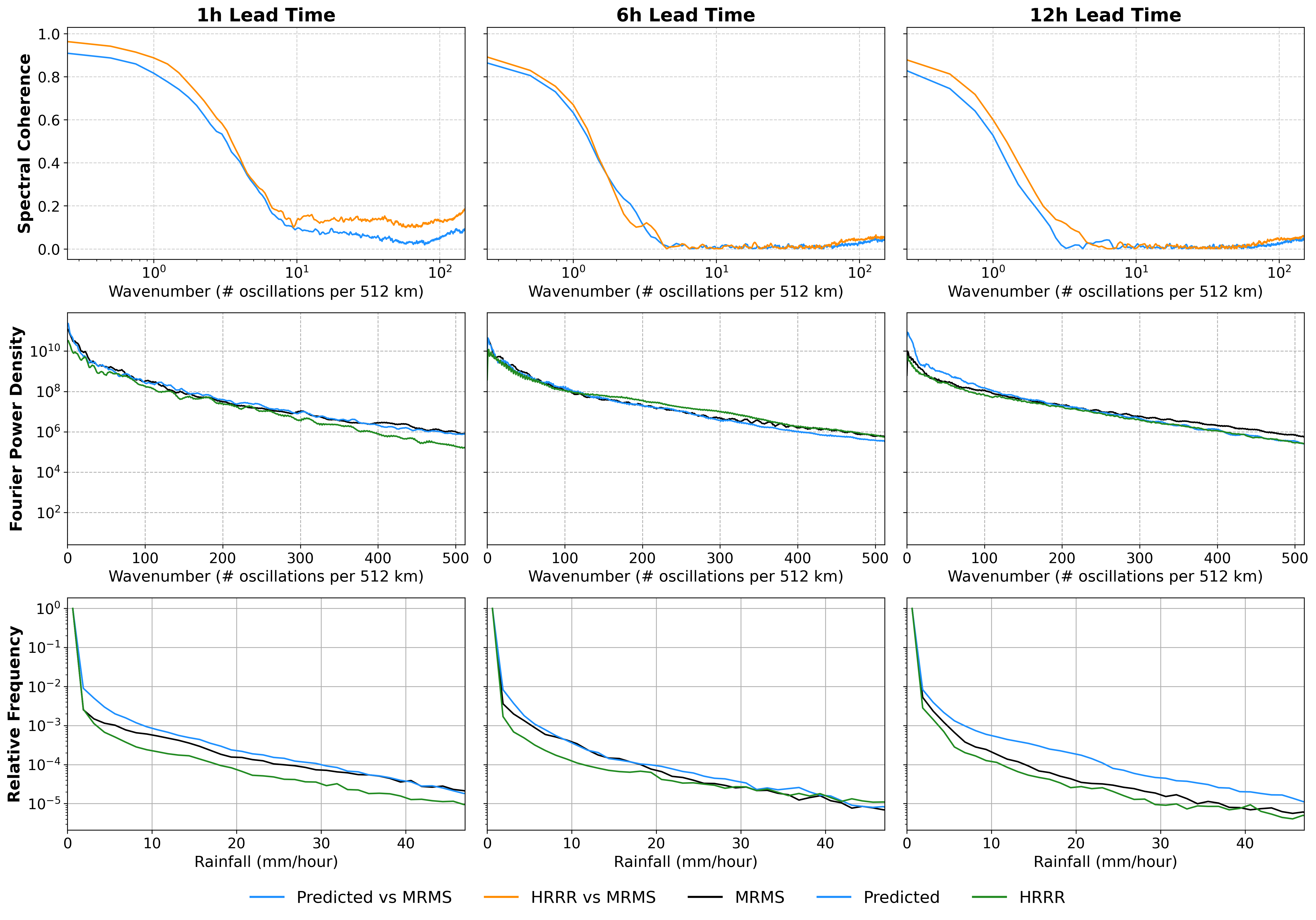}\\
\caption{Spatiostatistical evaluation of precipitation forecasts for lead times of 1h, 6h, and 12h during the median‑performing month (May). The first row compares the spectral coherence of the predicted fields with MRMS and of the HRRR forecasts with MRMS. The second row shows the Fourier Power Density Spectrum for MRMS, HRRR, and the predicted fields. The third row presents the rainfall intensity distribution as relative‑frequency histograms. Together, these metrics assess spatial structure, scale representation, and distributional fidelity across models and lead times.}
 \label{fig_spatiostatistic_metrics}
\end{center}
\end{figure}

To assess spatial realism, we evaluate the HRRR-Corrective model against the HRRR using three spatiostatistical metrics: spectral coherence, Fourier power density, and the rainfall intensity distribution. Results are shown in Figure~\ref{fig_spatiostatistic_metrics} for representative lead times (1h, 6h, 12h) using May 2024 as a median-performance month.

Spectral coherence quantifies the consistency of frequency components between model predictions and MRMS, reflecting the degree to which spatial rainfall patterns align across scales. While the HRRR-Corrective model shows slightly lower coherence than the HRRR—particularly at intermediate and high wavenumbers—this reduction likely arises from the residual learning process, which tends to smooth noisy small-scale features during correction. This behavior indicates that the corrective model prioritizes large-scale structural consistency and intensity realism over perfect pixel-wise phase alignment, where minor positional shifts or spatial smoothing can reduce coherence without degrading the overall predictive skill, especially when the corrected field better reproduces intensity and frequency statistics.

The 2D Fourier power spectrum further supports this interpretation. At 1h and 6h lead times, the HRRR-Corrective model exhibits power distributions that more closely match MRMS across a wide range of spatial frequencies, indicating improved scale representation. HRRR systematically underrepresents higher-frequency components, resulting in overly smooth precipitation fields. The corrective model mitigates this underestimation, preserving more realistic small-scale variability while maintaining the appropriate spectral slope. At 12h lead time, both models show attenuation at high frequencies, but the corrective model retains stronger power consistency with observations, suggesting more physically plausible storm texture and organization.

The rainfall intensity distribution, expressed as a relative-frequency histogram, reveals further improvements. The HRRR-Corrective model produces a probability density function that more closely aligns with MRMS, particularly in the moderate-to-heavy rainfall range (10–20~mm/h), where HRRR continues to underestimate event frequency. At longer lead times (e.g., 12h), the corrective model slightly overpredicts total rainfall but still maintains a more realistic distribution across intensities, reflecting enhanced calibration of rainfall magnitude despite phase or position errors.

Taken together, these metrics highlight complementary strengths. While the HRRR-Corrective model exhibits slightly lower spectral coherence—reflecting reduced pixel-level alignment—it achieves improved spatial power representation and rainfall intensity fidelity, supported by consistent gains in pixel-wise evaluation metrics (MAE, FSS, CSI). This indicates that residual correction enhances the physical realism of the rainfall field, favoring accurate scale and intensity reconstruction even when small-scale spatial phase alignment is less precise.

\subsection{Regional Evaluation}
\begin{figure}[t]
\begin{center}
 \noindent\includegraphics[width=\linewidth,angle=0]{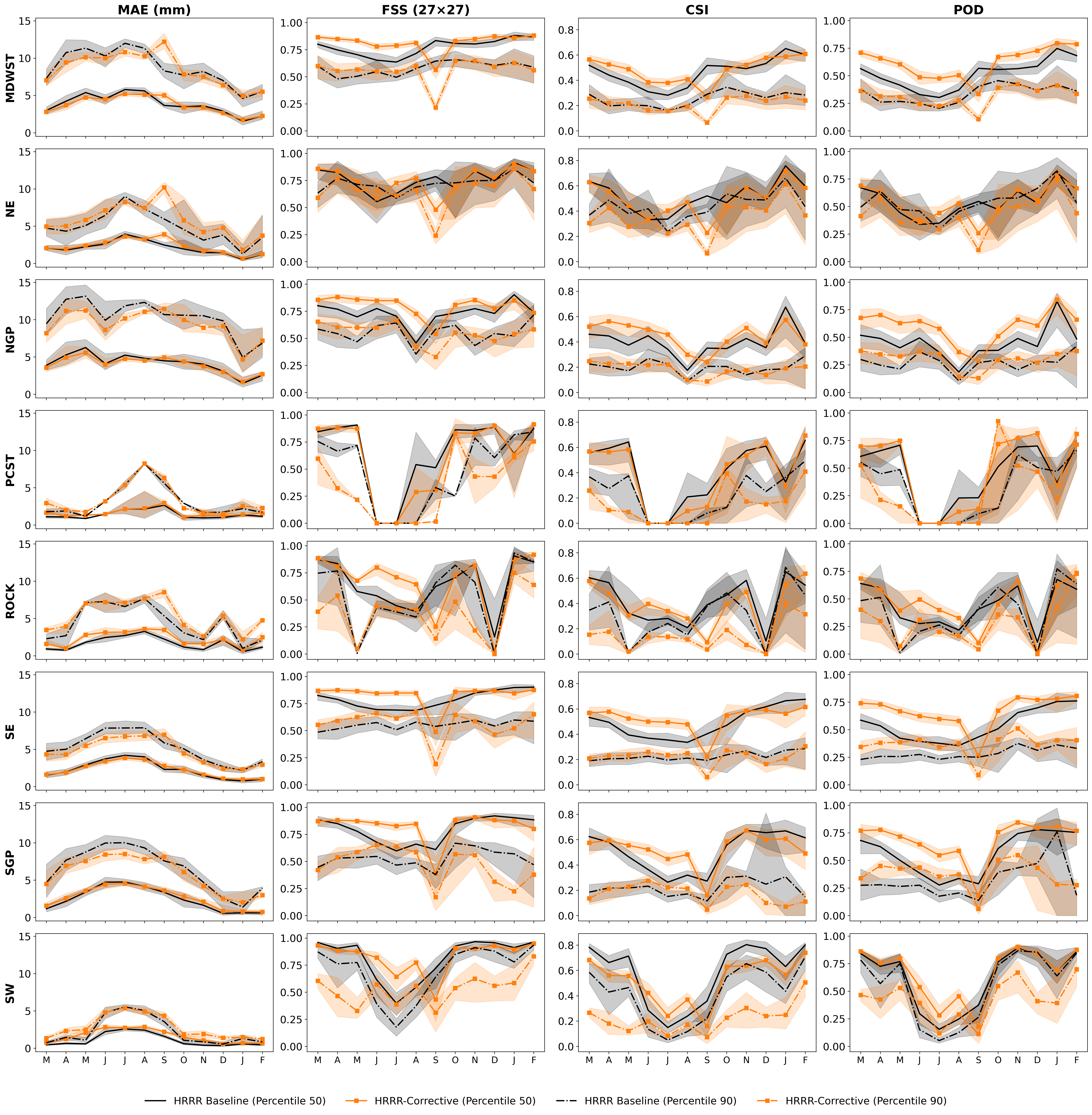}\\
\caption{Regional pixel-wise evaluation of 1h precipitation forecasts across the eight predefined CONUS regions (MDWST, NE, NGP, PCST, ROCK, SE, SGP, and SW), covering the period from March 2023 to February 2024. Each letter on the x-axis corresponds to the first letter of the month (e.g., M = March, A = April, etc.). Metrics shown include MAE, FSS (27×27), CSI, and POD. For each region, results are computed at both the 50th and 90th percentile rainfall thresholds to assess model skill for typical (median) precipitation as well as more extreme rainfall events. Comparisons are made between the HRRR and the HRRR-Corrective model.}
 \label{fig_regional_eval_1h}
\end{center}
\end{figure}

\begin{figure}[t]
\begin{center}
 \noindent\includegraphics[width=\linewidth,angle=0]{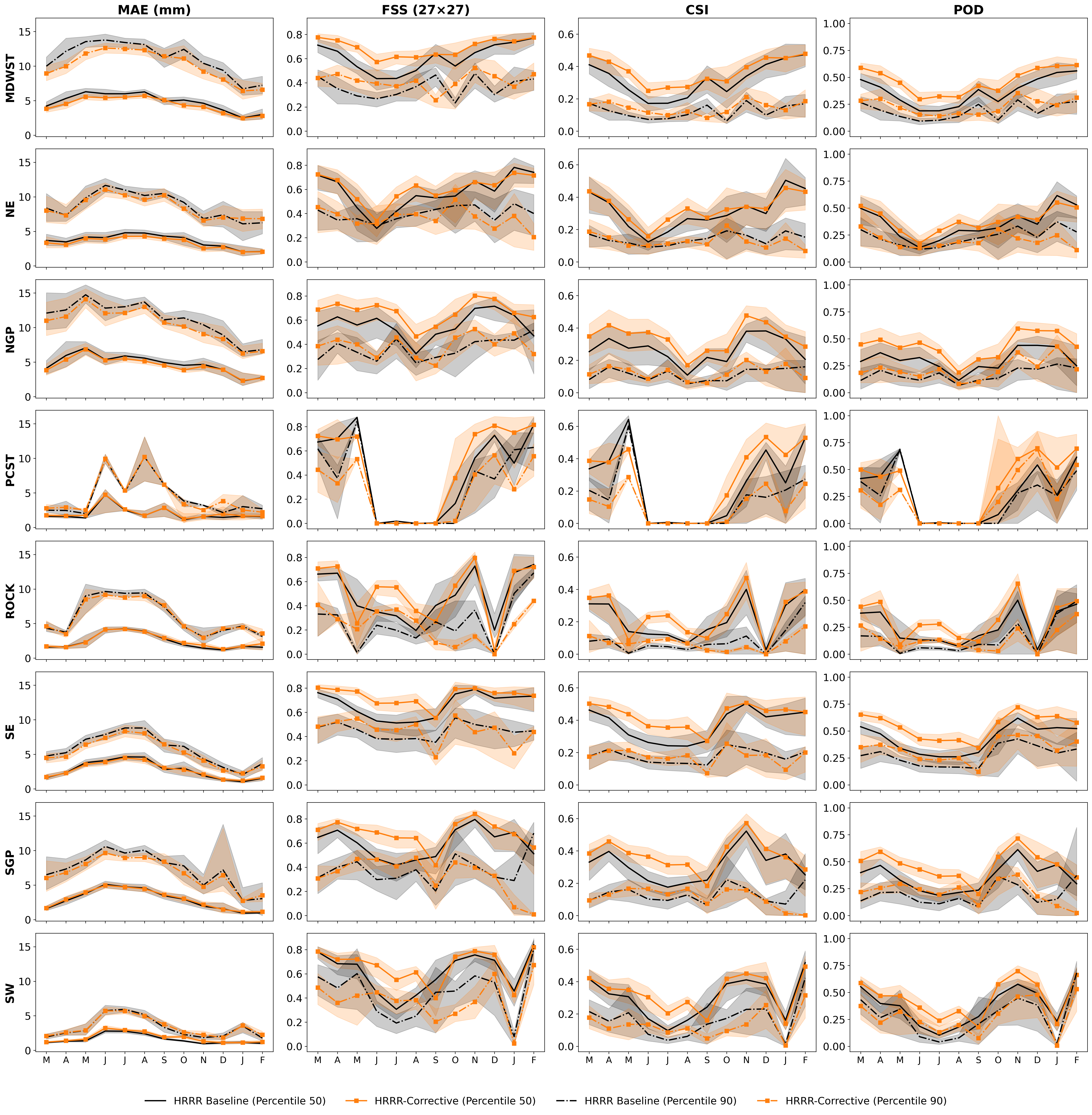}\\
\caption{Same as Figure~\ref{fig_regional_eval_1h}, but for 6h precipitation forecasts.}
 \label{fig_regional_eval_6h}
\end{center}
\end{figure}

To assess geographic robustness and temporal consistency, we evaluate the HRRR-Corrective model across eight CONUS subregions (Figure~\ref{fig_regions}) for both 1-hour and 6-hour forecasts. Figures~\ref{fig_regional_eval_1h} and~\ref{fig_regional_eval_6h} summarize monthly performance using MAE, FSS~(27×27), CSI, and POD metrics computed at the 50th and 90th percentile rainfall thresholds (see Table~\ref{tab:region_thresholds} in Appendix~\ref{append_evaluation_metrics}). These results enable assessment of model skill for both typical and extreme precipitation events under varying climatological and topographic regimes.

At the 1-hour lead time, performance patterns are generally consistent across regions, with the highest skill in the Southern Great Plains (SGP), Northern Great Plains (NGP), and Southeast (SE), where convective systems are frequent and well-resolved. In contrast, performance degrades in more complex regions such as the Rockies (ROCK) and Pacific Coast (PCST), where terrain and coastal influences introduce greater spatial variability. Notably, many regions show statistically significant improvements at the 50th percentile from March through August, particularly in the Midwest (MDWST), which exhibits the strongest overall performance across all metrics. Most regions also display a temporary decline in skill during September, likely reflecting seasonal transitions in storm regimes.

The 6-hour forecasts (Figure~\ref{fig_regional_eval_6h}) show overall higher errors and reduced skill compared to the 1-hour case, particularly for metrics sensitive to spatial organization (FSS, CSI). Nonetheless, the HRRR-Corrective model continues to outperform HRRR across most regions at the 50th percentile, with statistically significant improvements for MAE and FSS in many areas from March through August. These gains demonstrate that the corrective framework retains predictive value even at longer lead times. Error growth with lead time is most pronounced in convectively active regions such as the SGP and SE, underscoring the increasing difficulty of accurately representing storm initiation and propagation at multi-hour horizons.

Overall, the HRRR-Corrective model achieves its best skill in regions characterized by organized convection and flat terrain, while performance diminishes in areas with complex topography or coastal influences. The consistent month-to-month and multi-lead evaluation underscores the model’s geographic robustness and reveals systematic spatial and seasonal patterns in predictability, which are further explored in Section~\ref{visualization_section}.

\subsection{UQ Evaluation}
To evaluate the performance of the proposed UQ method—which provides both the middle prediction and associated lower and upper uncertainty bounds—we compute two complementary metrics: the coverage rate and the average error within the predicted interval.

The coverage rate measures the proportion of ground truth values that fall within the predicted uncertainty bounds. To accommodate spatial variability and potential displacement errors in high-resolution (1 km) forecasts, we apply a spatial tolerance of 10 km. Specifically, a prediction is considered “covered” if the ground truth value lies within the uncertainty bounds at any pixel within a 10 km radius. This adjustment accounts for the fact that strict pixel-level matching can penalize accurate predictions that are slightly displaced in space.

However, coverage alone may be misleading if the predicted intervals are too wide. Therefore, we also compute the average absolute error within the predicted interval, which quantifies the sharpness and informativeness of the UQ output. High coverage with low interval error indicates a well-calibrated and precise uncertainty estimate, while high coverage with large interval error would suggest overcautious predictions.

\begin{figure}[t]
    \centering
    \subfloat[1-hour]{\includegraphics[width=\textwidth]{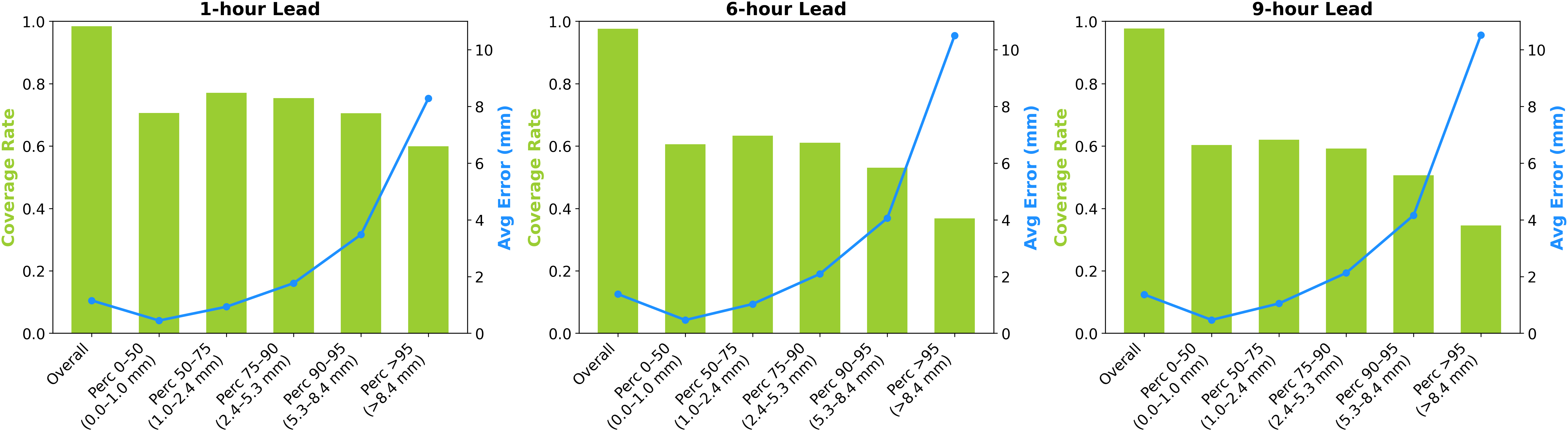}}\hfill
    \caption{UQ evaluation of HRRR-corrective forecasts. Subplots (a) and (b) show the coverage rate (green bars, left axis) and average error (blue line, right axis) across precipitation intensity bins for the 1-hour, and 6-hour, and 9-hour lead time predictions, respectively.}
    \label{fig_uq}
\end{figure}

Figure~\ref{fig_uq} summarizes the UQ performance across precipitation intensity bins for 1-hour, 6-hour, and 9-hour lead time predictions. The model demonstrates stable and reliable UQ, with coverage rates exceeding 50\% across nearly all intensity bins and lead times—except beyond the 95th percentile, where uncertainty naturally increases due to the rarity and variability of extreme rainfall, even under strict binning and exclusion of zero values. 

For high-intensity rainfall, particularly in the 95th percentile bin, the average error within the predicted interval remains around 8~mm for the 1-hour forecast and slightly above 10~mm for the 6- and 9-hour forecasts. This indicates that while coverage is lower for the most extreme events, the model’s misses are relatively small—highlighting the robustness and reliability of the predicted intervals. Further analysis shows a modest performance drop between the direct 1-hour predictions and autoregressive rollouts, which is expected; however, beyond 6 hours, the coverage remains stable, supporting the temporal consistency of the UQ mechanism.

Additionally, similar coverage is observed across percentiles up to the 90th, with a noticeable decline only beyond that point. This demonstrates that the model effectively captures uncertainty across the full range of moderate to heavy rainfall events, where the average error remains below or equal to 4~mm for all lead times.

Overall, these results show that the proposed UQ framework provides consistent and well-calibrated coverage across both intensity ranges and time horizons, while maintaining narrow and informative uncertainty intervals. This balance is particularly valuable for forecasting rare and high-impact precipitation events, where both precision and reliability are critical.

\subsection{Prediction Visualization}\label{visualization_section}
\begin{figure}[h]
    \centering
    \subfloat[]{\includegraphics[width=\textwidth]{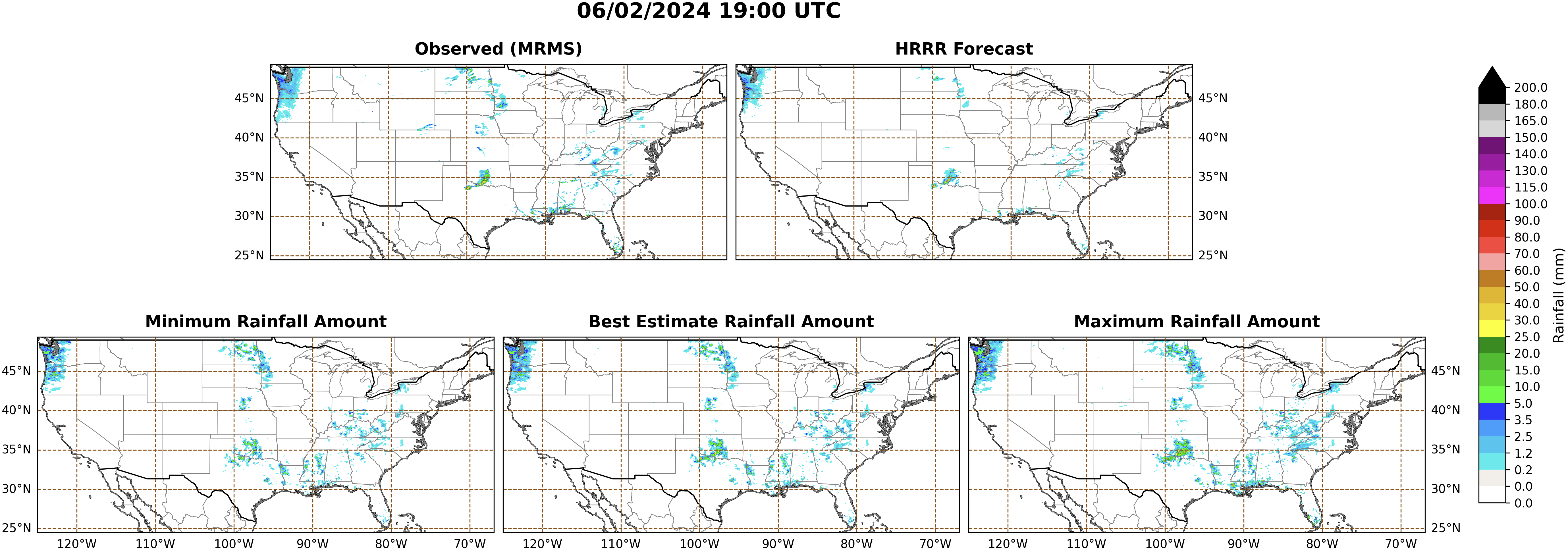}}\\
    \subfloat[]{\includegraphics[width=\textwidth]{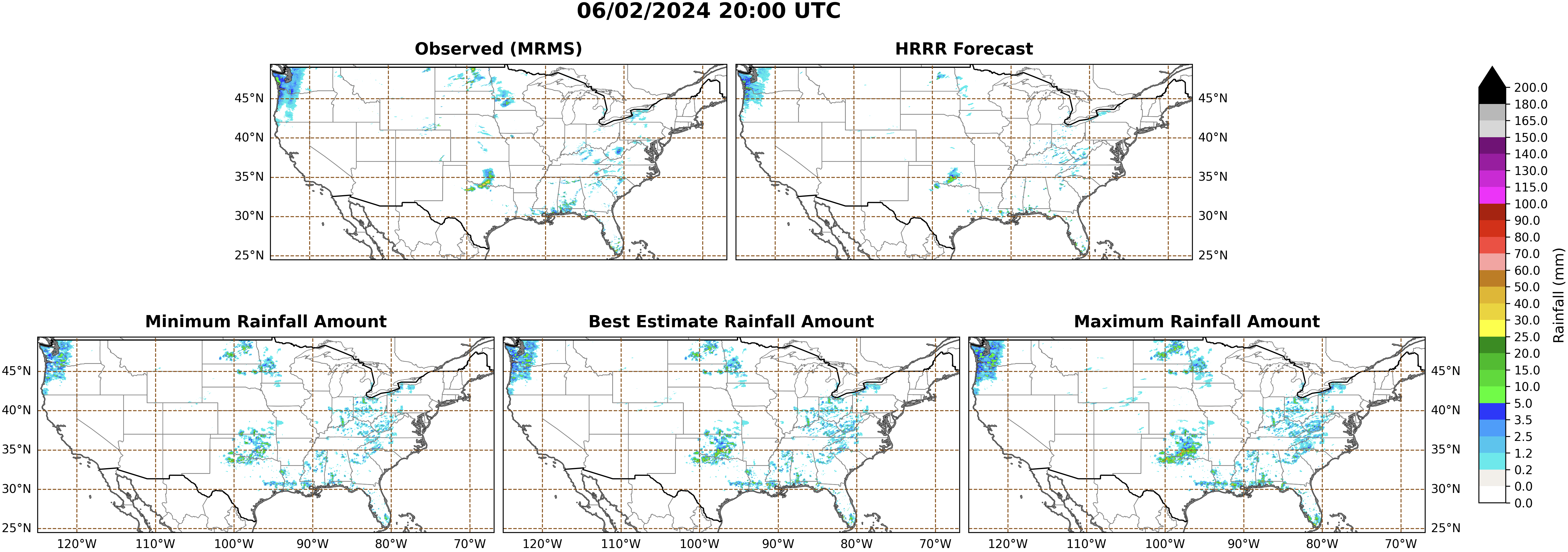}}\\
    \subfloat[]{\includegraphics[width=\textwidth]{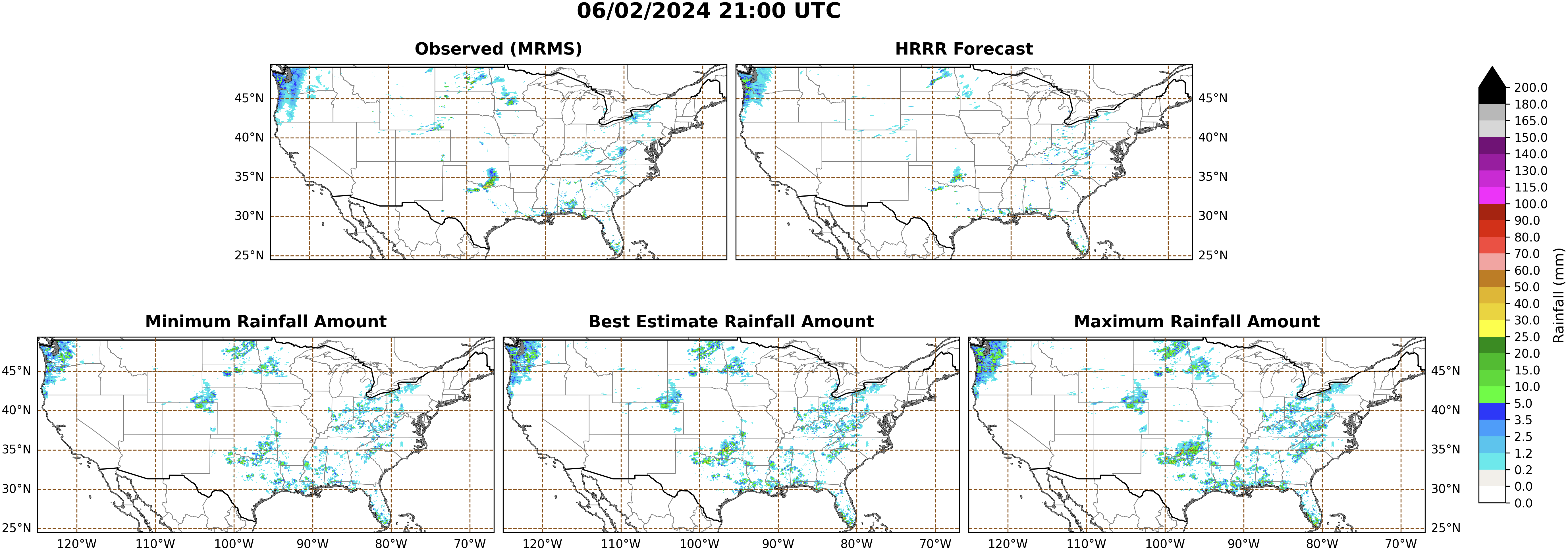}}
    \caption{Corrective HRRR forecasts for a case scenario on June 2, 2024. Subplots (a), (b), and (c) show the 1-hour, 2-hour, and 3-hour predictions, respectively.}
    \label{fig_pred_with_uq}
\end{figure}

Figure~\ref{fig_pred_with_uq} illustrates a representative example of the HRRR‑Corrective model’s predictive capability and associated UQ. The figure compares MRMS observations, HRRR forecasts, and the corrective model outputs at 1-, 2-, and 3-hour lead times, showing minimum, middle, and maximum rainfall predictions that together represent the predicted uncertainty interval.

This example highlights the advantage of the proposed UQ approach. Unlike traditional ensemble-based uncertainty estimation, which is computationally expensive in our diffusion framework, our method derives uncertainty directly from the residual-based reconstruction of precipitation fields, combining multiple HRRR forecasts to quantify prediction variability. This enables efficient spatiotemporal characterization of confidence levels while maintaining high resolution. When comparing MRMS, HRRR, and the corrective model, we observe that the HRRR-Corrective configuration better captures small-scale precipitation features and enhanced rainfall magnitudes—particularly in convective regions where HRRR tends to underdetect precipitation—aligning more closely with MRMS observations.

Second, the evolution from the 1h to 3h demonstrates that the model tracks the progression of the storm with notable spatial and structural consistency. Importantly, the UQ bounds remain relatively narrow across all lead times, indicating that the model does not inflate uncertainty as the forecast horizon increases. This reflects a stable and well‑calibrated UQ mechanism whose intervals remain informative while avoiding excessive spread.

\begin{figure}[htbp]
    \centering
    \subfloat[]{\includegraphics[width=\textwidth]{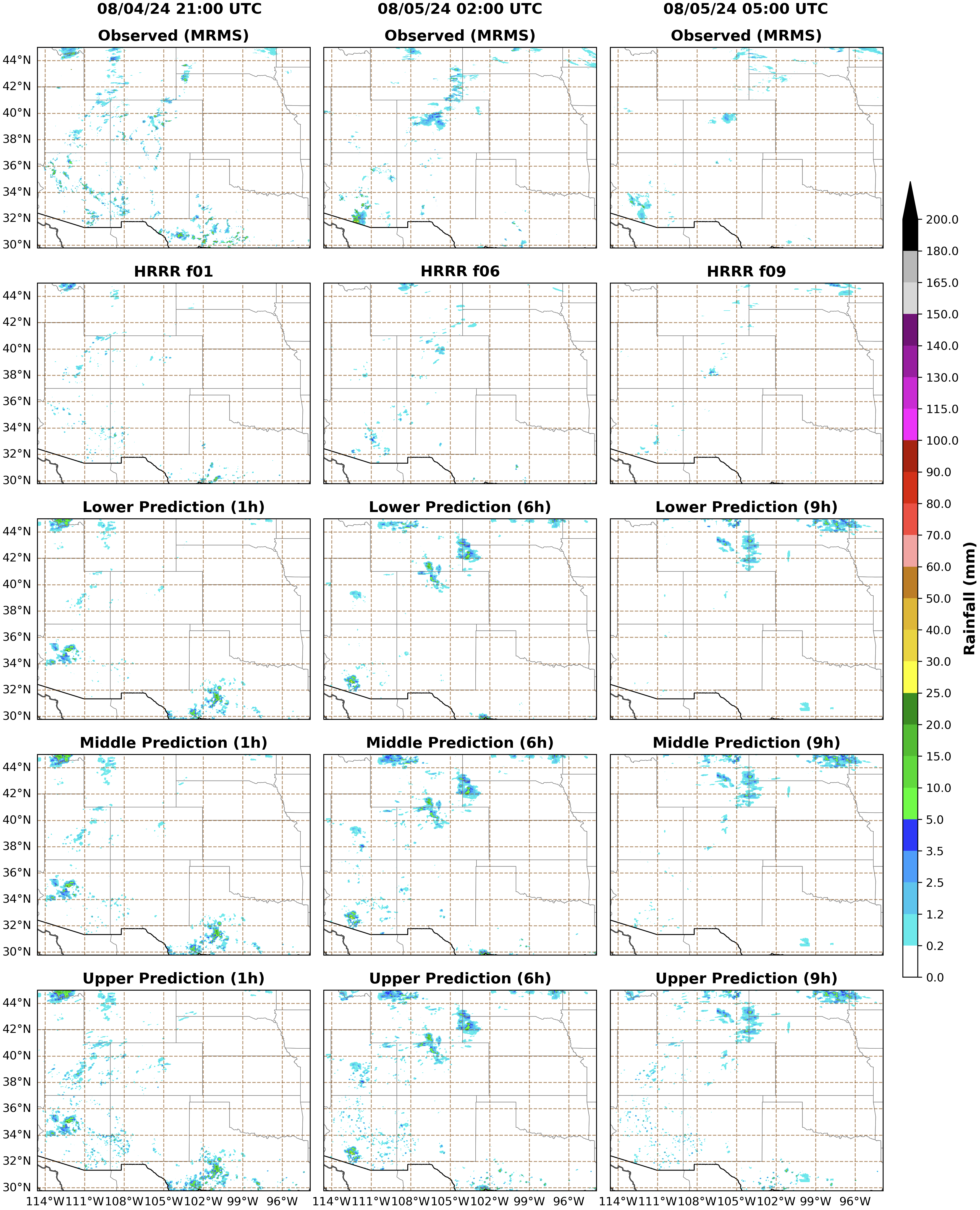}}\\
    \caption{Comparison between the corrective HRRR model, the HRRR and observations for three forecast lead times (1 h, 6 h, 9 h; columns). The top row shows the observed rainfall from MRMS at the corresponding valid times (indicated above each column). The second row presents the raw HRRR forecasts (f01, f06, f09). The lower three rows show the lower, middle, and upper predictive predictions from the corrective model, representing the uncertainty range of the AI predictions.}
    \label{fig_zoomed_pred_with_uq}
\end{figure}

To further compare the HRRR-Corrective model against HRRR and the MRMS observations, we refer to Figure~\ref{fig_zoomed_pred_with_uq}, which provides a zoomed-in view to better visualize differences among the models. This example clearly illustrates one of the main advantages of the HRRR-Corrective approach—its ability to enhance predictions of small-scale rainfall features, a task where HRRR often underperforms. Additionally, the figure highlights the benefit of incorporating UQ through the lower, middle, and upper predictive bounds. The upper-bound predictions ensure that fine-scale, high-intensity rainfall is captured, while the lower-bound predictions help prevent overestimation, together providing a balanced and interpretable representation of forecast uncertainty.

Together, these qualitative results complement the quantitative evaluations, reinforcing that the HRRR‑Corrective approach not only improves over the HRRR in detecting and representing fine‑scale rainfall features, but also provides reliable and sharp uncertainty estimates throughout the short‑term forecast window.

\subsection{Comparison with the State of the Art}
The work most closely related to this study is the MetNet trilogy \citep{MetNet1,MetNet2,MetNet3}, which represents the only family of deep learning models that provide CONUS-wide precipitation forecasts at high spatial resolution and longer lead times—up to 8~h for MetNet-1, 12~h for MetNet-2, and 24~h for MetNet-3. In contrast, most other ML and DL approaches focus on short-term nowcasting (typically 0--6~h) and are often limited to localized domains rather than generalizing across the entire CONUS \citep{Gao2023,Asperti2025,Chase2025}. Moreover, many of these models predict binary outcomes (e.g., rain/no-rain or threshold exceedance), while our framework produces rainfall magnitude predictions capable of accurately representing both light and extreme precipitation regimes.

Despite their conceptual similarity, there are several notable differences between our framework and MetNet-3. First, MetNet-3 generates probabilistic, categorical forecasts, predicting the probability that rainfall will fall within predefined intensity bins. In contrast, our model directly predicts continuous rainfall magnitudes and provides UQ in the form of upper and lower confidence bounds. While probabilistic predictions offer valuable information, our approach provides an alternative, computationally efficient representation of forecast uncertainty that can be directly visualized as confidence intervals—potentially more intuitive for certain applications. This distinction is particularly important for forecasting extreme rainfall, where accurate magnitude estimation is more actionable than probabilistic exceedance information.

A second distinction lies in the number of input variables and computational requirements. MetNet-3 leverages hundreds of meteorological fields and extensive computational infrastructure, reportedly using 512~TPUv3 cores for seven days of training. In contrast, our models attain high predictive skill using a concise input set—primarily past MRMS observations and HRRR precipitation forecasts—trained for 16, 21, and 8~days on only two NVIDIA~H100 or four A100 GPUs. This efficiency stems from the models’ ability to learn and exploit the inherent spatiotemporal correlations and persistence in HRRR and MRMS fields, which provide the dominant sources of predictability in short-term rainfall forecasting. Although our training durations are longer, the hardware requirements are substantially reduced, improving reproducibility and accessibility for research environments. Furthermore, our framework bridges data-driven learning and physically informed modeling by explicitly incorporating HRRR forecasts as inputs, yielding greater interpretability and physical consistency than purely statistical approaches.

Another key contribution of this work lies in the systematic comparison of three distinct modeling strategies—data-driven, HRRR-corrective, and hybrid—each leveraging different sources of information. To our knowledge, no previous study has explicitly contrasted these approaches within a unified framework. This comparison not only advances DL precipitation forecasting but also deepens our understanding of the physical processes underlying precipitation predictability. Our results demonstrate that purely data-driven models hold strong potential for very short lead times, while hybrid approaches that integrate numerical forecasts with observations are essential for longer lead times, when new storm systems develop and observational persistence alone becomes insufficient.

Furthermore, this work introduces, for the first time in this field, a comprehensive evaluation framework that assesses model skill across both spatial and temporal dimensions. Specifically, we perform regional and monthly evaluations and report performance using both pixel-wise and spatiostatistical metrics. This unified framework provides a more complete and nuanced characterization of forecast skill than has been available in previous studies.

Overall, both MetNet and our diffusion-based framework represent complementary advances in AI-driven precipitation forecasting. Whereas MetNet emphasizes probabilistic event likelihoods, our model focuses on magnitude predictions with physically informed uncertainty quantification—two perspectives that together expand the frontier of DL weather prediction.

\subsection{Model Limitations}
A common limitation of many AI-based correction models, including ours, is that although they effectively learn and compensate for the systematic biases present in numerical models such as HRRR, they may introduce new, smaller biases of their own. In our case, the HRRR-Corrective model exhibits mild overprediction tendencies in specific situations—particularly during spring months in the Southeast and during summer over the Rockies. These regional and seasonal discrepancies suggest that certain physical processes or atmospheric conditions may be inherently more difficult for the model to correct or learn from. Importantly, despite these localized biases, the HRRR-Corrective model consistently outperforms the original HRRR forecasts across most regions, seasons, lead times, and evaluation metrics. This indicates that the introduced biases are relatively minor compared to the larger error patterns the model successfully mitigates.

Future work could investigate whether these residual biases stem from unresolved or poorly represented physical processes in the HRRR model itself (e.g., convective initiation, orographic lifting, or moisture transport). This could be examined by comparing AI-model errors with HRRR physics tendencies or reforecast experiments that isolate individual parameterization impacts, helping to identify whether systematic biases originate from the HRRR forcing fields. These effects may be especially influential during months and regions where the DL model shows reduced performance.

\section{Conclusion and Future Work}
This manuscript presents a comprehensive evaluation of three AI-based precipitation forecasting models across the CONUS: a data-driven model, a hybrid model, and an HRRR-corrective model. These models are benchmarked against the operational HRRR forecasts to assess their performance at both regular and extreme rainfall intensities, with 1 km spatial resolution and lead times ranging from 1 to 12 hours. A central novelty of this work lies in the direct comparison across three DL models each with distinct advantages and limitations. To the best of our knowledge, this is the first study to conduct such a large-scale and structured comparison for CONUS-wide extreme precipitation, providing critical insight into how different inputs in DL models enhance learning under real-world operational constraints. Our results demonstrate that the hybrid model achieves the best performance at short lead times (1h), while the HRRR-corrective model consistently outperforms both the HRRR and other DL models at longer lead times (up to 12h). This indicates the value of incorporating physics-based forecasts as a conditional prior when predicting complex spatiotemporal rainfall patterns.

Another contribution of this work is the detailed monthly and regional evaluation of model performance. We show that the DL model behaviors are not only statistically robust but also physically consistent with regional precipitation regimes. For example, model skill drops in regions and months characterized by convective or monsoonal activity, which are historically more challenging to predict. This physics-informed analysis framework adds interpretability and reliability to AI-based weather forecasting.

We also introduce a UQ pipeline with enhanced interpretability for operational use. A new custom loss function is developed to handle spatial error structures and false positive/negative asymmetries in rainfall prediction. Furthermore, we propose a novel UQ framework that quantifies and visualizes spatiotemporal uncertainty to support stakeholder decision-making, bridging the gap between raw model outputs and actionable insights.

Future work will explore the development of a globally generalizable DL model for rainfall prediction, with scalability to diverse climates and input data sources. We also aim to refine our UQ framework with conformal prediction, which provides statistically valid uncertainty intervals for each forecast. Additionally, we plan to enhance model interpretability by incorporating explainability metrics (e.g., saliency maps, input-attribution analyses) to identify key predictors influencing rainfall intensity. These improvements will make the model’s outputs more transparent and actionable for end users such as emergency managers and forecasters.

%

%

\clearpage
\acknowledgments
This material is based upon work supported by the National Science Foundation under Grant No. RISE-2019758 within the NSF AI Institute for Research on Trustworthy AI in Weather, Climate, and Coastal Oceanography (AI2ES), and under Grant No. IIS2324008. Any opinions, findings, and conclusions or recommendations expressed in this material are those of the author(s) and do not necessarily reflect the views of the National Science Foundation. 

%
%
\datastatement
The datasets used in this study are publicly available through NOAA’s Amazon Web Server (AWS) Open Data Program. The MRMS precipitation data can be accessed at https://registry.opendata.aws/noaa-mrms-pds/, and the HRRR forecast data are available at https://registry.opendata.aws/noaa-hrrr-pds/. The code developed for model training, and evaluation will be made publicly available on GitHub upon publication of this work. 

%

\appendix[A]
\section{Input Variable Evaluation and Ablation Studies}
\label{append_inputs_evaluated}

To determine the most informative set of predictors for high-resolution precipitation forecasting, we conducted comprehensive ablation studies evaluating a broad range of HRRR forecast variables alongside MRMS observations. These experiments aimed to assess the marginal utility of each candidate input in order to identify a configuration that balances predictive performance, generalization, and computational efficiency.

\subsection{Evaluated Input Variables}

Table~\ref{tab_inputs_evaluated} lists the HRRR forecast fields considered during our ablation studies, grouped by physical category. These include thermodynamic, dynamic, moisture-related, and other environmental variables commonly used in meteorological modeling. Multiple forecast lead times are tested for each variable to account for short-term atmospheric evolution and uncertainty.

\begin{table}[t]
\caption{HRRR forecast fields evaluated during ablation studies, grouped by variable type. Only the most important variables, as identified through these studies, are used in the final model.}
\label{tab_inputs_evaluated}
\centering
\resizebox{\textwidth}{!}{
\begin{tabular}{cccc}
\hline\hline
\textit{Thermodynamics Fields} & \textit{Dynamic Fields} & \textit{Moisture \& Precipitation} & \textit{Others} \\
\hline
Temperature at 500 hPa & Vertical Vorticity at 2000 m AGL & Specific Humidity at 2 m & Terrain elevation \\
Geopotential Height at 500 hPa & Vertical Velocity at 500 hPa & Dewpoint Temperature at 2 m & \\
Mean Sea Level Pressure & U \& V Wind at 2000 m AGL and 10 m & Precipitable Water & \\
Convective Available Potential Energy & U \& V Wind at 500 hPa & Total Precipitation (1-h Accumulation) & \\
Convective Inhibition & & Composite Reflectivity & \\
& & Reflectivity at 1000 m AGL & \\
\hline
\end{tabular}
}
\end{table}

\subsection{Ablation Results and Final Input Selection}
While several of the evaluated variables—such as Convective Available Potential Energy (CAPE), precipitable water, and reflectivity—are physically relevant and have been used in prior forecasting studies, our ablation results showed that their inclusion did not improve model performance. In many cases, they degraded performance due to issues such as input redundancy, noise, or inconsistent spatiotemporal patterns across forecast cycles. This outcome aligns with established challenges in deep learning, where the inclusion of excessive or highly correlated variables can introduce conflicting signals, increase overfitting risk, and reduce model generalization. Our findings emphasize that a parsimonious input set—grounded in physical relevance and aligned with the task at hand—can outperform more complex or exhaustive input configurations.

In addition to forecast fields, we tested the use of auxiliary contextual inputs. Latitude and longitude coordinates are included to provide spatial awareness, while cyclic temporal encodings (hour of day, day of year, and month) enabled the model to capture diurnal and seasonal patterns. These inputs consistently improved performance, likely because they help the model account for the varying physical processes and seasonal dynamics that characterize different regions across the CONUS.

Lastly, we assess the value of including prior MRMS QPE observations at multiple timesteps. These past observations contributed significantly to short-term forecasting performance, particularly for 1-hour lead time predictions, consistent with broader findings in the environmental modeling literature. This temporal context proves more valuable than many of the additional HRRR forecast fields tested.

\appendix[B]
\section{Tile Sampling Strategy for Training}
\label{appendix_tile_sampling}
To construct a diverse and informative training dataset, we follow a multi-step tile sampling and filtering strategy designed to maximize spatial, temporal, and rainfall variability, while ensuring regional balance across CONUS.

\subsection{Step 1: Raw Tile Sampling Based on Rainfall Criteria}
We begin by randomly sampling 512×512 km tiles from across the CONUS at each timestep. A tile is retained if it meets at least one of the following two conditions:
\begin{itemize}
    \item Rainfall coverage: At least 25\% of pixels contain non-zero precipitation.
    \item Extreme rainfall: At least one pixel exceeds the local 10-year Annual Recurrence Interval (ARI), as estimated by NOAA Atlas 14 \citep{NOAA_Atlas14_PFDS}.
\end{itemize}

To prevent spatial redundancy and the overrepresentation of localized convective systems, we enforce a minimum spacing of 30 km between selected tiles. For each timestep, up to 50 candidate tiles are evaluated until 20 valid ones are retained. If fewer than 20 are found, the subset of valid tiles is used and we proceed to the next timestep. This sampling is performed independently for each month in the training set.

The output of this step is a large intermediate dataset containing hundreds of thousands of tile-date pairs, reflecting a wide range of precipitation events, spatial distributions, and atmospheric conditions.

\subsection{Step 2: Regionally Balanced Subsampling}
To prevent geographic bias and ensure that all climatic regimes are equally represented during training, we subsample from the large pool generated in Step~1 by selecting up to 120 valid tiles per region per month. Regions follow the NOAA-defined hydrometeorological boundaries (e.g., SE, NGP, PCST). If a region–month pair contains fewer than 120 tiles, all available samples are retained. This strategy avoids overrepresentation of wetter regions while preserving coverage across drier areas, resulting in a dataset that is both spatially diverse and event-rich.

\appendix[C]
\section{Evaluation Sampling Strategy}
\label{appendix_testing_strategy}
For each month in the testing period (March~2024 to February~2025), we select two forecast initialization times to evaluate model performance. These times are determined using a shifting scheme that varies by both month to ensure broad diurnal and seasonal coverage without clustering evaluations at the same hours.

The specific selection follows a modular arithmetic procedure. For a given month string (e.g., "2024\_03"), we extract the numeric month and year, compute a base offset, and apply a month shift to obtain a unique pair of evaluation times. The two resulting hours are spaced 12 hours apart (e.g., 3Z and 15Z), yielding balanced sampling between daytime and nighttime conditions. This strategy ensures that every month is evaluated under distinct and representative atmospheric conditions while keeping the number of predictions tractable.

\appendix[D]
\section{Hyperparameters and Computational Resources}
\label{append_hyperparams}
This appendix summarizes the key training configurations and computational resources used in this study. These details complement the methodological descriptions in Section~\ref{sec_methodology} and are provided here for reproducibility and completeness.

The data-driven and hybrid models are trained on two NVIDIA H100 GPUs (80~GB each), while the HRRR-Corrective model is trained on four NVIDIA A100 GPUs (80~GB each). Total training time varied significantly: approximately 16 days for the data-driven model, 21 days for the hybrid model, and 8 days for the HRRR-Corrective model.

Due to these high computational demands—particularly the 21-day runtime for the hybrid setup—full-scale hyperparameter tuning over the entire dataset is not feasible. Instead, tuning is conducted on a representative data subset containing approximately 10\% of the total 30,000 samples, selected to preserve the spatial and temporal diversity of the full CONUS dataset and to identify robust configurations transferable to large-scale training.

All models are implemented in TensorFlow 2.17 with CUDA version 12.4 and trained using float32 precision. The final configuration used the Adam optimizer with a learning rate of $1 \times 10^{-5}$ and a batch size of 2. While larger batches are often recommended for diffusion models due to improved gradient stability, we observe better performance with smaller batches. This is likely because over 75\% of pixels in the dataset correspond to zero rainfall, and larger batches dilute the frequency of informative events. Early stopping is applied with a patience of 100 epochs and a minimum delta of $1 \times 10^{-6}$, restoring the best-performing weights based on training loss.

Note that predictions are generated over tiles of size $512 \times 512$~pixels with a 50~km overlap between adjacent tiles. As a result, 78 tile-level predictions are required to reconstruct the full CONUS domain. To determine the most effective strategy for merging overlapping regions, we conduct several experiments comparing different aggregation schemes, including minimum, maximum, average, and weighted combinations of the overlapping pixels. The minimum and maximum methods led to degraded performance and produced unphysical edge artifacts, while the average yielded the most consistent and accurate reconstructions. Therefore, the averaging approach is adopted for all final CONUS-scale products.

\appendix[E]
\section{Weighted MAE Weighting Curve}
\label{appendix_weighted_mae_curve}
As described in Section~\ref{sec_methodology}\ref{sec_loss}, the HybridSigmaLoss combines two complementary components: a Scaled~MAE term that incorporates predictive uncertainty, and a Weighted~MAE term that adjusts the loss according to rainfall intensity. This appendix focuses specifically on the latter component to illustrate how the weighting mechanism behaves across the full precipitation range.

\begin{figure}[h!]
    \centering
    \includegraphics[width=0.6\textwidth]{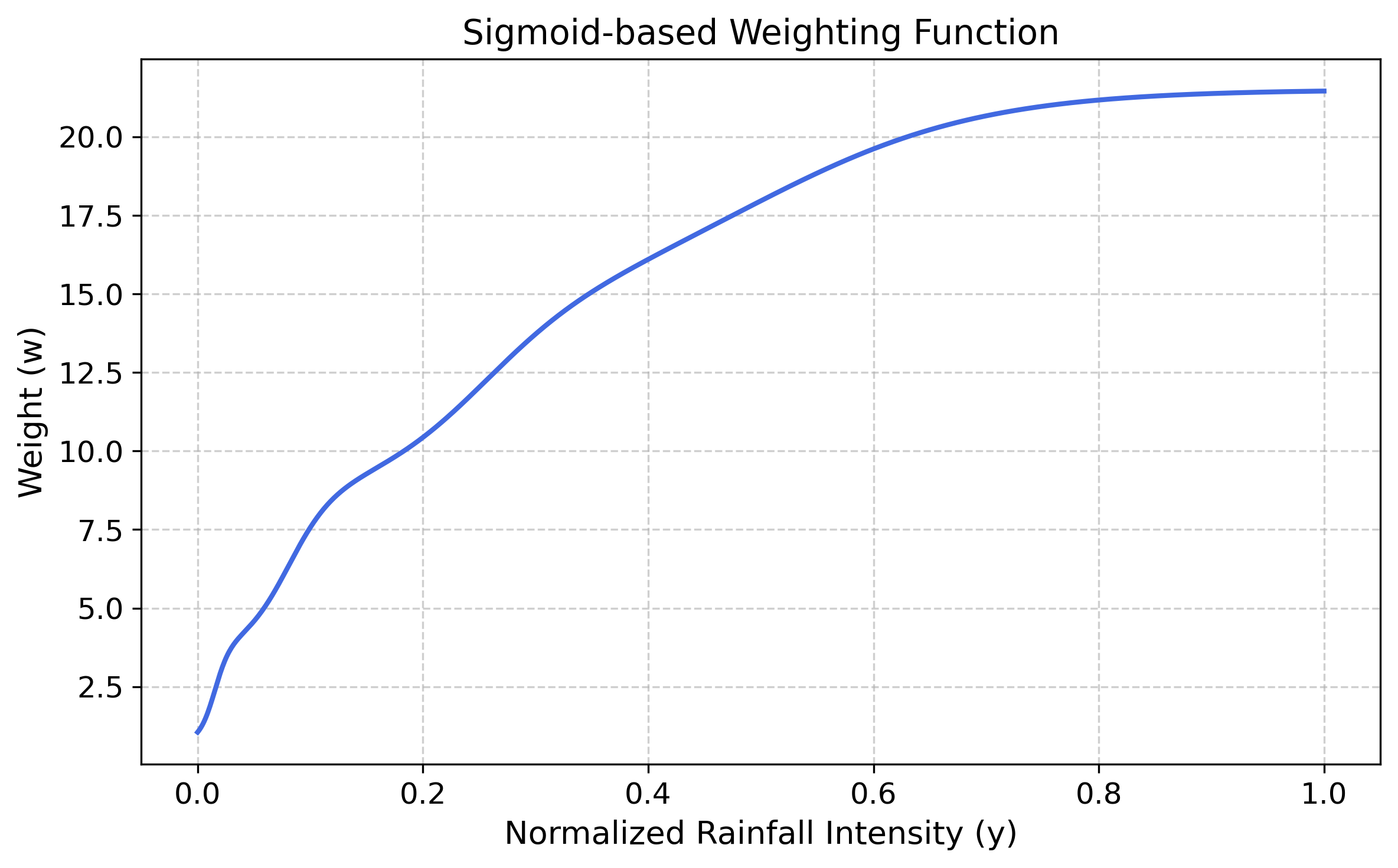}
    \caption{Sigmoid-based weighting function used in the Weighted MAE component of the HybridSigmaLoss. Weights increase with rainfall intensity, producing a smooth transition between low- and high-penalty regions.}
    \label{fig_weight_curve}
\end{figure}

Figure~\ref{fig_weight_curve} shows the sigmoid-based weighting function defined in Equation~\ref{eq_weighted_mae_weights}. The curve progressively increases the penalty with rainfall intensity, allowing the model to emphasize rare, high-impact precipitation events while maintaining smooth transitions and numerical stability during optimization. This design enables the loss to remain responsive to both light- and heavy-rain regimes.

\appendix[F]
\section{Evaluation Metrics Details}
\label{append_evaluation_metrics}
This appendix provides the definitions and mathematical formulations of the evaluation metrics used throughout the manuscript. Metrics are grouped into two complementary categories: (i) pixel‑wise and threshold‑based metrics that quantify pointwise accuracy and event detection, and (ii) spatiostatistical metrics that assess structural realism and spatial coherence with the true precipitation field (MRMS). Together, these metrics offer a multifaceted view of model performance relative to MRMS ground truth.

\subsection{Pixel-Wise and Threshold-Based Metrics}
Pixel-wise metrics evaluate performance at the grid-cell level and form the basis for assessing accuracy and classification skill across different rainfall intensities.

\paragraph{Mean Absolute Error (MAE)}  
\begin{equation}
\mathrm{MAE} = \frac{1}{N}\sum_{i=1}^{N} \left| \hat{y}_i - y_i \right|
\label{eq_mae}
\end{equation}
where \( N \) is the total number of pixels, \( \hat{y}_i \) is the predicted rainfall at pixel \( i \), and \( y_i \) is the corresponding MRMS value.

\paragraph{Probability of Detection (POD)}
\begin{equation}
\mathrm{POD} = \frac{TP}{TP + FN}
\label{eq_pod}
\end{equation}
where \( TP \) is the number of true positives and \( FN \) is the number of false negatives.

\paragraph{Critical Success Index (CSI)}
\begin{equation}
\mathrm{CSI} = \frac{TP}{TP + FP + FN}
\label{eq_csi}
\end{equation}

Where \( FP \) denotes false positives.

\paragraph{Fraction Skill Score (FSS)}  
\begin{equation}
\mathrm{FSS} = 1 - \frac{\mathrm{MSE}(f, o)}{\mathrm{MSE}_{\mathrm{ref}}}
\label{eq_fss}
\end{equation}

Here, \( f \) and \( o \) represent the fractional coverage of predicted and observed rainfall events over a neighborhood (5 by 5), \( \mathrm{MSE}(f, o) \) is the mean squared error between them, and \( \mathrm{MSE}_{\mathrm{ref}} \) is the MSE of a random reference forecast.

Thresholds used for binary metrics vary by region; see Table~\ref{tab:region_thresholds} for the full regional threshold list.

\begin{table}[ht]
\centering
\caption{Rainfall thresholds (in mm) used for binary evaluation metrics across regions.}
\label{tab:region_thresholds}
\begin{tabular}{cccccc}
\toprule
\textbf{Region} & \textbf{50th Perc.} & \textbf{75th Perc.} & \textbf{90th Perc.} & \textbf{95th Perc.} \\
\midrule
\textbf{CONUS}  & 1.02 & 2.40 & 5.26 & 8.43 \\
\textbf{PCST}   & 0.96 & 1.89 & 3.18 & 4.21 \\
\textbf{ROCK}   & 0.90 & 1.92 & 3.94 & 6.13 \\
\textbf{NGP}    & 1.14 & 2.80 & 6.33 & 10.14 \\
\textbf{MDWST}  & 1.33 & 3.15 & 6.66 & 10.27 \\
\textbf{NE}     & 0.88 & 2.03 & 4.71 & 7.72 \\
\textbf{SW}     & 0.72 & 1.41 & 2.52 & 3.55 \\
\textbf{SGP}    & 0.99 & 2.17 & 4.47 & 7.04 \\
\textbf{SE}     & 0.96 & 2.04 & 4.00 & 5.90 \\
\bottomrule
\end{tabular}
\end{table}

\subsection{Spatiostatistical Metrics}
Spatiostatistical metrics evaluate whether the predicted precipitation fields reproduce the spatial structure, frequency content, and distributional properties of MRMS.

\paragraph{2D Fourier Power Spectrum}
\begin{equation}
P(k_x, k_y) = \left| \hat{Y}(k_x, k_y) \right|^2
\label{eq_2d_fourier}
\end{equation}
where \( \hat{Y}(k_x, k_y) \) is the 2D discrete Fourier transform of the predicted precipitation field, and \( P(k_x, k_y) \) is the spectral power at wavenumbers \( k_x \) and \( k_y \), corresponding to the horizontal and vertical spatial directions, respectively.

\paragraph{Spectral Coherence}
\begin{equation}
\mathrm{Coherence}(f) = 
\frac{ \left| S_{xy}(f) \right| }
     { \sqrt{ S_{xx}(f) \, S_{yy}(f) } }
\label{eq_spec_coher}
\end{equation}
where \( S_{xy}(f) \) is the cross-spectral density of the predicted and observed fields at frequency \( f \), and \( S_{xx}(f) \), \( S_{yy}(f) \) are the power spectral densities of the prediction and observation, respectively.

\paragraph{Rainfall Intensity Distribution (PDF)}
\begin{equation}
\mathrm{PDF}(b_j) = \frac{1}{N}
\sum_{i=1}^{N} \mathbf{1}_{b_j}(y_i)
\label{eq_pdf}
\end{equation}
where \( b_j \) denotes the \( j^{th} \) histogram bin, and \( \mathbf{1}_{b_j}(y_i) \) is an indicator function equal to 1 if the MRMS rainfall value \( y_i \) falls in bin \( b_j \), and 0 otherwise.



%



\bibliographystyle{ametsocV6}
\bibliography{references}

@STRING{AN        = "Astrophys.\ Norv."}

@STRING{AO        = "Atmos.--Ocean"}

@STRING{MA        = "Meteor.\ Appl."}

@article{Hill_Schumacher2021,
  author = {A. Hill and R. S. Schumacher},
  title = {Forecasting excessive rainfall with random forests and a deterministic convection-allowing model},
  journal = {Wea. Forecasting},
  year = {2021},
  volume = {36},
  number = {5},
  pages = {1693--1711},
}

@article{HoEA2020,
  author = {J. Ho and A. Jain and P. Abbeel},
  title = {Denoising diffusion probabilistic models},
  journal = {Adv. Neural Inf. Process. Syst.},
  year = {2020},
  volume = {33},
  pages = {6840--6851}
}

@article{KarrasEA2022,
  author = {T. Karras and M. Aittala and T. Aila and S. Laine},
  title = {Elucidating the design space of diffusion-based generative models},
  journal = {Adv. Neural Inf. Process. Syst.},
  year = {2022},
  volume = {35},
  pages = {26565--26577}
}

@article{SongEA2020,
  author = {Y. Song and J. Sohl-Dickstein and D. P. Kingma and A. Kumar and S. Ermon and B. Poole},
  title = {Score-based generative modeling through stochastic differential equations},
  journal = {arXiv preprint},
  year = {2020},
  eprint = {arXiv:2011.13456},
  archivePrefix = {arXiv}
}

@article{Beniston2013,
  author = {Beniston, Martin},
  title = {Grand challenges in climate research},
  journal = {Frontiers in Environmental Science},
  year = {2013},
  volume = {1},
  pages = {1},
}

@techreport{Alexander2016,
  author = {Alexander, L. V. and Zhang, X. and Hegerl, G. and Seneviratne, S. I. and Behrangi, A. and Fischer, E. and Vautard, R. and others},
  title = {Implementation plan for WCRP grand challenge on understanding and predicting weather and climate extremes -- the ``Extremes Grand Challenge''},
  institution = {World Climate Research Programme},
  year = {2016},
  note = {Version: June 2016},
  url = {https://www.wcrp-climate.org/images/documents/grand_challenges/GC_Extremes_v2.pdf}
}

@techreport{NOAA2020,
  author       = {{National Oceanic and Atmospheric Administration}},
  title        = {{Precipitation Prediction Grand Challenge}},
  year         = {2020},
  institution  = {U.S. Department of Commerce},
  note         = {Available online at \url{https://www.noaa.gov/sites/default/files/2022-01/PPGC-Strategy_FINAL_2020-1030.pdf}}
}

@incollection{Smith1979,
  author = {Smith, R. B.},
  title = {The influence of mountains on the atmosphere},
  booktitle = {Advances in Geophysics},
  volume = {21},
  pages = {87--230},
  year = {1979},
  publisher = {Elsevier},
  editor = {Landsberg, H. E. and Van Mieghem, J.}
}

@article{Veneziano2006,
  author = {Veneziano, D. and Langousis, A. and Furcolo, P.},
  title = {Multifractality and rainfall extremes: A review},
  journal = {Water Resources Research},
  volume = {42},
  number = {6},
  year = {2006},
}

@article{Tabari2020,
  author = {Tabari, H.},
  title = {Climate change impact on flood and extreme precipitation increases with water availability},
  journal = {Scientific Reports},
  volume = {10},
  number = {1},
  pages = {13768},
  year = {2020},
}

@misc{HRRRweb,
  author = {{NOAA Earth System Research Laboratory}},
  title = {{High-Resolution Rapid Refresh (HRRR)}},
  howpublished = {\url{https://rapidrefresh.noaa.gov/hrrr/}},
  note = {Accessed September 2024},
  year = {n.d.}
}

@inproceedings{Sangiorgio2019,
  author = {Sangiorgio, M. and Barindelli, S. and Biondi, R. and Solazzo, E. and Realini, E. and Venuti, G. and Guariso, G.},
  title = {Improved extreme rainfall events forecasting using neural networks and water vapor measures},
  booktitle = {International Conference on Time Series and Forecasting - Proceedings of Papers - Volumen 2},
  pages = {820--826},
  year = {2019}
}

@inproceedings{Gope2016,
  author = {Gope, S. and Sarkar, S. and Mitra, P. and Ghosh, S.},
  title = {Early prediction of extreme rainfall events: a deep learning approach},
  booktitle = {Advances in Data Mining. Applications and Theoretical Aspects: 16th Industrial Conference, ICDM 2016, New York, NY, USA, July 13--17, 2016},
  publisher = {Springer International Publishing},
  year = {2016}
}

@article{Chkeir2023,
  author = {Chkeir, S. and Anesiadou, A. and Mascitelli, A. and Biondi, R.},
  title = {Nowcasting extreme rain and extreme wind speed with machine learning techniques applied to different input datasets},
  journal = {Atmospheric Research},
  volume = {282},
  pages = {106548},
  year = {2023}
}

@article{Vitanza2023,
  author = {Vitanza, E. and Dimitri, G. M. and Mocenni, C.},
  title = {A multi-modal machine learning approach to detect extreme rainfall events in Sicily},
  journal = {Scientific Reports},
  volume = {13},
  number = {1},
  pages = {6196},
  year = {2023}
}

@article{Kagabo2024,
  author = {Kagabo, J. and Kattel, G. R. and Kazora, J. and Shangwe, C. N. and Habiyakare, F.},
  title = {Application of Machine Learning Algorithms in Predicting Extreme Rainfall Events in Rwanda},
  journal = {Atmosphere},
  volume = {15},
  number = {6},
  pages = {691},
  year = {2024}
}

@misc{Chase2025,
  author = {Chase, R. J. and Haynes, K. and Hoef, L. V. and Ebert-Uphoff, I.},
  title = {Score-based diffusion nowcasting of GOES imagery},
  howpublished = {arXiv preprint arXiv:2505.10432},
  year = {2025}
}

@misc{MetNet1,
  author = {Sønderby, C. K. and Espeholt, L. and Heek, J. and Dehghani, M. and Oliver, A. and Salimans, T. and Kalchbrenner, N.},
  title = {MetNet: A neural weather model for precipitation forecasting},
  howpublished = {arXiv preprint arXiv:2003.12140},
  year = {2020}
}

@article{MetNet2,
  author = {Espeholt, L. and Agrawal, S. and Sønderby, C. and Kumar, M. and Heek, J. and Bromberg, C. and Kalchbrenner, N.},
  title = {Deep learning for twelve hour precipitation forecasts},
  journal = {Nature Communications},
  volume = {13},
  number = {1},
  pages = {5145},
  year = {2022}
}

@misc{MetNet3,
  author = {Andrychowicz, M. and Espeholt, L. and Li, D. and Merchant, S. and Merose, A. and Zyda, F. and Kalchbrenner, N.},
  title = {Deep learning for day forecasts from sparse observations},
  howpublished = {arXiv preprint arXiv:2306.06079},
  year = {2023}
}

@article{Guilloteau2025,
  author = {Guilloteau, C. and Kerrigan, G. and Nelson, K. and Migliorini, G. and Smyth, P. and Li, R. and Foufoula-Georgiou, E.},
  title = {A generative diffusion model for probabilistic ensembles of precipitation maps conditioned on multisensor satellite observations},
  journal = {IEEE Transactions on Geoscience and Remote Sensing},
  year = {2025}
}

@inproceedings{Gao2023,
  author = {Gao, Z. and Shi, X. and Han, B. and Wang, H. and Jin, X. and Maddix, D. and Wang, Y. B.},
  title = {Prediff: Precipitation nowcasting with latent diffusion models},
  booktitle = {Advances in Neural Information Processing Systems},
  volume = {36},
  pages = {78621--78656},
  year = {2023}
}

@article{Asperti2025,
  author = {Asperti, A. and Merizzi, F. and Paparella, A. and Pedrazzi, G. and Angelinelli, M. and Colamonaco, S.},
  title = {Precipitation nowcasting with generative diffusion models},
  journal = {Applied Intelligence},
  volume = {55},
  number = {3},
  pages = {187},
  year = {2025}
}

@article{Benjamin2016,
  author = {Benjamin, S. G. and Weygandt, S. S. and Brown, J. M. and Hu, M. and Alexander, C. R. and Smirnova, T. G. and Olson, J. B. and James, E. P. and Dowell, D. C. and Grell, G. A. and Lin, H. and Peckham, S. E. and Smith, T. L. and Moninger, W. R. and Kenyon, J. S. and Manikin, G. S.},
  title = {A North American Hourly Assimilation and Model Forecast Cycle: The Rapid Refresh},
  journal = {Monthly Weather Review},
  volume = {144},
  number = {4},
  pages = {1669--1694},
  year = {2016},
}

@article{Dougherty2020,
  author = {Dougherty, E. and Rasmussen, K. L.},
  title = {Changes in Future Flash Flood--Producing Storms in the United States},
  journal = {Journal of Hydrometeorology},
  year = {2020},
  volume = {21},
  number = {10},
  pages = {2221--2236},
}

@article{Rasp2018,
  author = {Rasp, S. and Pritchard, M. S. and Gentine, P.},
  title = {Deep learning to represent subgrid processes in climate models},
  journal = {Proceedings of the National Academy of Sciences},
  volume = {115},
  number = {39},
  pages = {9684--9689},
  year = {2018},
}

@article{Dowell2022,
  author    = {Dowell, David C. and Alexander, Curtis R. and James, Eric P. and Weygandt, Stephen S. and Benjamin, Stanley G. and Manikin, George S. and Alcott, Timothy I. and others},
  title     = {The High‐Resolution Rapid Refresh (HRRR): An Hourly Updating Convection‐Allowing Forecast Model. Part I: Motivation and System Description},
  journal   = {Weather and Forecasting},
  year      = {2022},
  volume    = {37},
  number    = {8},
  pages     = {1371--1395},
}

@article{Li2012,
  author = {Li, F. and Rosa, D. and Collins, W. D. and Wehner, M. F.},
  title = {“Super‐parameterization”: A Better Way to Simulate Regional Extreme Precipitation?},
  journal = {Journal of Advances in Modeling Earth Systems},
  volume = {4},
  number = {2},
  pages = {M04002},
  year = {2012},
}

@misc{Zhang2014,
  author = {Zhang, X. and Hegerl, G. and Seneviratne, S. and Stewart, R. and Zwiers, F. and Alexander, L.},
  title = {WCRP Grand Challenge: Understanding and Predicting Weather and Climate Extremes},
  year = {2014},
  howpublished = {\url{https://www.wcrp-climate.org/grand-challenges/grand-challenges-overview}},
  note = {WCRP White Paper}
}

@book{NRC2006,
  title = {Completing the Forecast: Characterizing and Communicating Uncertainty for Better Decisions Using Weather and Climate Forecasts},
  author = {National Research Council},
  year = {2006},
  publisher = {The National Academies Press},
  address = {Washington, DC},
}

@article{Loken2019,
  author = {Loken, Eric D. and Clark, Adam J. and McGovern, Amy and Flora, Matthew and Knopfmeier, Karen},
  title = {Postprocessing Next-Day Ensemble Probabilistic Precipitation Forecasts Using Random Forests},
  journal = {Weather and Forecasting},
  volume = {34},
  number = {6},
  pages = {2017--2044},
  year = {2019},
}

@phdthesis{JamesDissertation,
  author = {Eric James},
  title = {Comparing Precipitation Estimates, Model Forecasts, and Random Forest Based Predictions for Excessive Rainfall},
  school = {Colorado State University},
  year = {2023},
  note = {Ph.D. dissertation},
  url = {https://www.proquest.com/openview/9398e158d428cf33e2d82912669cbf05/1?pq-origsite=gscholar&cbl=18750&diss=y&casa_token=_urU8QS5rpQAAAAA:nYyPV7_Hs6xbHHb9ZfgqHRfSK0B0YHm6WKlxE1Ru8oqcSbZdtpiTGhbNvBLFoDL1uh4glavd8g}
}

@article{Zhang2016,
  author = {J. Zhang and K. Howard and C. Langston and B. Kaney and Y. Qi and L. Tang and D. Kitzmiller and S. Stevens and S. Dunning and L. Wang and Z. Li and B. Martinaitis and J. Cocks and K. Vasiloff and C. Calhoun and J. Hall and G. Stumpf and D. Miller and R. Demo},
  title = {Multi-Radar Multi-Sensor (MRMS) quantitative precipitation estimation: Initial operating capabilities},
  journal = {Bulletin of the American Meteorological Society},
  volume = {97},
  number = {4},
  pages = {621--638},
  year = {2016},
}

@misc{NOAA_Atlas14_PFDS,
  author = {{NOAA National Weather Service}},
  title = {Precipitation Frequency Data Server (PFDS) – NOAA Atlas 14},
  year = {n.d.},
  howpublished = {\url{https://hdsc.nws.noaa.gov/hdsc/pfds/}},
  note = {Accessed September 2024}
}

@article{English2021,
  author  = {English, J. M. and Turner, D. D. and Alcott, T. I. and Moninger, W. R. and Bytheway, J. L. and Cifelli, R. and Marquis, M.},
  title   = {Evaluating Operational and Experimental HRRR Model Forecasts of Atmospheric River Events in California},
  journal = {Weather and Forecasting},
  year    = {2021},
  volume  = {36},
  number  = {6},
  pages   = {1925--1944},
}

@inproceedings{Shi2015ConvLSTM,
  author    = {Xingjian Shi and Zhourong Chen and Hao Wang and Dit-Yan Yeung and Wai-Kin Wong and Wang-chun Woo},
  title     = {Convolutional {LSTM} Network: A Machine Learning Approach for Precipitation Nowcasting},
  booktitle = {Advances in Neural Information Processing Systems (NeurIPS)},
  volume    = {28},
  year      = {2015},
}

@inproceedings{yu2024unmasking,
  title={Unmasking bias in diffusion model training},
  author={Yu, Hu and Shen, Li and Huang, Jie and Li, Hongsheng and Zhao, Feng},
  booktitle={European Conference on Computer Vision},
  pages={374--390},
  year={2024},
  organization={Springer}
}

@article{meijer2024rise,
  title={The rise of diffusion models in time-series forecasting},
  author={Meijer, Caspar and Chen, Lydia Y},
  journal={arXiv preprint arXiv:2401.03006},
  year={2024}
}

@misc{Hwang2024,
  title={Increasing dam failure risk in the USA due to compound rainfall clusters as climate changes. npj Natural Hazards, 1 (1), 27},
  author={Hwang, J and Lall, U},
  year={2024}
}

@article{Dharmarathne2024,
  title={Adapting cities to the surge: A comprehensive review of climate-induced urban flooding},
  author={Dharmarathne, Gangani and Waduge, AO and Bogahawaththa, Madhusha and Rathnayake, Upaka and Meddage, DPP},
  journal={Results in Engineering},
  volume={22},
  pages={102123},
  year={2024},
  publisher={Elsevier}
}

@article{heinselman2024warn,
  title={Warn-on-Forecast System: From vision to reality},
  author={Heinselman, Pamela L and Burke, Patrick C and Wicker, Louis J and Clark, Adam J and Kain, John S and Gao, Jidong and Yussouf, Nusrat and Jones, Thomas A and Skinner, Patrick S and Potvin, Corey K and others},
  journal={Weather and Forecasting},
  volume={39},
  number={1},
  pages={75--95},
  year={2024},
  publisher={American Meteorological Society}
}

@article{walser2004predictability,
  title={Predictability of precipitation in a cloud-resolving model},
  author={Walser, Andr{\'e} and L{\"u}thi, Daniel and Sch{\"a}r, Christoph},
  journal={Monthly Weather Review},
  volume={132},
  number={2},
  pages={560--577},
  year={2004}
}

@article{jeworrek2021wrf,
  title={WRF precipitation performance and predictability for systematically varied parameterizations over complex terrain},
  author={Jeworrek, Julia and West, Gregory and Stull, Roland},
  journal={Weather and Forecasting},
  volume={36},
  number={3},
  pages={893--913},
  year={2021}
}

@article{li2024statistical,
  title={Statistical characteristics of convective and stratiform precipitation during the rainy season over South China based on GPM-DPR observations},
  author={Li, Donghuan and Qi, Youcun and Li, Huiqi},
  journal={Atmospheric Research},
  volume={301},
  pages={107267},
  year={2024},
  publisher={Elsevier}
}

@article{balmaseda2021noaa,
  title={NOAA-DOE Precipitation Processes and Predictability Workshop Report},
  author={Balmaseda, Magdalena and Barros, Ana and Hagos, Samson and Kirtman, Ben and Ma, Hsi-Yen and Ming, Yi and Pendergrass, Angie and Tallapragada, Vijay and Thompson, Elizabeth},
  year={2021}
}

@article{skamarock2008description,
  title={A description of the advanced research WRF version 3, NCAR Technical Note},
  author={Skamarock, William C and Klemp, Joseph B and Dudhia, Jimy and Gill, David O and Barker, Dale M and Duda, Michael G and Huang, Xiang-Yu and Wang, Wei and Powers, Jordan G},
  journal={National Center for Atmospheric Research: Boulder, CO, USA},
  pages={113},
  year={2008}
}

@article{merz2020impact,
  title={Impact forecasting to support emergency management of natural hazards},
  author={Merz, Bruno and Kuhlicke, Christian and Kunz, Michael and Pittore, Massimiliano and Babeyko, Andrey and Bresch, David N and Domeisen, Daniela IV and Feser, Frauke and Koszalka, Inga and Kreibich, Heidi and others},
  journal={Reviews of geophysics},
  volume={58},
  number={4},
  pages={e2020RG000704},
  year={2020},
  publisher={Wiley Online Library}
}

@article{turner2022evaluating,
  title={Evaluating the economic impacts of improvements to the High-Resolution Rapid Refresh (HRRR) numerical weather prediction model},
  author={Turner, David D and Cutler, Harvey and Shields, Martin and Hill, Rebecca and Hartman, Brad and Hu, Yuchen and Lu, Tao and Jeon, Hwayoung},
  journal={Bulletin of the American Meteorological Society},
  volume={103},
  number={2},
  pages={E198--E211},
  year={2022}
}

@article{mardani2025residual,
  title={Residual corrective diffusion modeling for km-scale atmospheric downscaling},
  author={Mardani, Morteza and Brenowitz, Noah and Cohen, Yair and Pathak, Jaideep and Chen, Chieh-Yu and Liu, Cheng-Chin and Vahdat, Arash and Nabian, Mohammad Amin and Ge, Tao and Subramaniam, Akshay and others},
  journal={Communications Earth \& Environment},
  volume={6},
  number={1},
  pages={124},
  year={2025},
  publisher={Nature Publishing Group UK London}
}

@software{xESMF_zhuang_2018_1134366,
  author       = {Jiawei Zhuang},
  title        = {JiaweiZhuang/xESMF: v0.1.1},
  month        = jan,
  year         = 2018,
  publisher    = {Zenodo},
  version      = {v0.1.1},
  doi          = {10.5281/zenodo.1134366},
  url          = {https://doi.org/10.5281/zenodo.1134366},
}

@article{xarray_hoyer2017,
  title={xarray: ND labeled arrays and datasets in Python},
  author={Hoyer, Stephan and Hamman, Joe},
  journal={Journal of open research software},
  volume={5},
  number={1},
  pages={10--10},
  year={2017}
}

@article{kain2008some,
  title={Some practical considerations regarding horizontal resolution in the first generation of operational convection-allowing NWP},
  author={Kain, John S and Weiss, Steven J and Bright, David R and Baldwin, Michael E and Levit, Jason J and Carbin, Gregory W and Schwartz, Craig S and Weisman, Morris L and Droegemeier, Kelvin K and Weber, Daniel B and others},
  journal={Weather and Forecasting},
  volume={23},
  number={5},
  pages={931--952},
  year={2008}
}

@article{weisman2023simulations,
  title={Simulations of severe convective systems using 1-versus 3-km grid spacing},
  author={Weisman, Morris L and Manning, Kevin W and Sobash, Ryan A and Schwartz, Craig S},
  journal={Weather and Forecasting},
  volume={38},
  number={3},
  pages={401--423},
  year={2023}
}

\end{document}